\documentclass[acmsmall,screen]{acmart}

\AtBeginDocument{%
  \providecommand\BibTeX{{%
    \normalfont B\kern-0.5em{\scshape i\kern-0.25em b}\kern-0.8em\TeX}}}

\setcopyright{acmcopyright}
\copyrightyear{2022}
\acmYear{2022}
\acmDOI{10.1145/3572843}

\acmJournal{TOMM}
\acmVolume{37}
\acmNumber{4}
\acmArticle{100}
\acmMonth{8}

\begin{document}

\title{Attention, Please! Adversarial Defense via Activation Rectification and Preservation}

\author{Shangxi Wu}
\email{wushangxi@bjtu.edu.cn}
\author{Jitao Sang}
\author{Kaiyuan Xu}
\author{Jiaming Zhang}
\author{Jian Yu}

\affiliation{
  \institution{Beijing Key Lab of Traffic Data Analysis and Mining, Beijing Jiaotong University}
  \city{Beijing}
  \country{China}
}

\renewcommand{\shortauthors}{Shangxi Wu, et al.}

\begin{abstract}
  This study provides a new understanding of the adversarial attack problem by examining the correlation between adversarial attack and visual attention change. In particular, we observed that: (1) images with incomplete attention regions are more vulnerable to adversarial attacks; and (2) successful adversarial attacks lead to deviated and scattered activation map. Therefore, we use the mask method to design an attention-preserving loss and a contrast method to design a loss that makes the model's attention rectification. Accordingly, an attention-based adversarial defense framework is designed, under which better adversarial training or stronger adversarial attacks can be performed through the above constraints. We hope the attention-related data analysis and defense solution in this study will shed some light on the mechanism behind the adversarial attack and also facilitate future adversarial defense/attack model design.
\end{abstract}


\maketitle

\section{Introduction}
Many standard image classification models are recognized to be highly vulnerable to adversarial attack, which adds small 
perturbation to the original samples but maliciously misleads the model prediction. Extensive studies have 
been conducted on designing different adversarial attack methods to fool state-of-the-art convolutional 
networks~\cite{szegedy2013intriguing,kurakin2016adversarial,Seyed-Mohsen2016Deepfool,Ian2017Explaining,nicholas2017towards, 7018027}. 
Applying adversarial attacks in automatic visual systems like self-driving vehicles can lead to catastrophic 
consequences~\cite{kurakin2016adversarial}, and it is thus necessary to develop effective defense methods against 
potential attacks~\cite{8649865, 8576563}.

\begin{figure}[t]
  \begin{center}
     \includegraphics[width=0.45\linewidth]{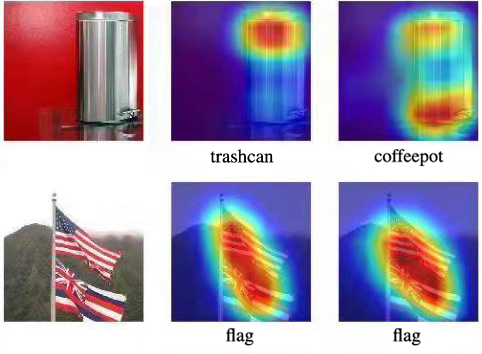}
  \end{center}
     \caption{Visual attention of the successfully-attacked image (top row) and the failed-attacked image (bottom row). In each row, we show the original image, the original image's activation map, and the adversarial image's activation map.}
  \label{fig:1}
\end{figure}

Adversarial training is recognized as a way of regularization and updating the decision boundary around adversarial 
samples~\cite{tramer2017ensemble}.
Adversarial training provides a fundamental and flexible defense framework compatible with different realizations. 
The performance of the specific realization depends on three factors: (1) Regularizing the model 
to focus on robust features. It is observed that adversarial perturbations amplify the importance of 
low-contribution features to change the output prediction~\cite{tsipras2018there}. Modifying models to restrict the 
model prediction focusing on robust features is expected to improve the robustness of original samples~\cite{DBLP:journals/tomccap/DuanLOWLT20}. (2) Reducing 
feature distribution divergence. In addition to injecting the adversarial samples into the training set, further constraints 
can be designed between original and adversarial samples to alleviate feature fluctuation caused by adversarial 
input perturbations. 

This study falls into the adversarial training group and attempts to address the above two factors to improve defense performance. Visual attention\footnote{In this paper, we define the activation map as visual attention.} has been used to explain which region of the image is responsible for the network's decision~\cite{zeiler2014visualizing}. Through data analysis in Sec.~\ref{section2}, we observed significant correlations between visual attention and adversarial attack. Fig.~\ref{fig:1} shows two example images from ImageNet2012. The activation map of the original images and the corresponding FGSM-created adversarial images toward the true labels are illustrated for comparison. Quick observations include: (1) By comparing the activation maps of the two original images, we found that the upper image relies on the fractional object region for prediction and turns out vulnerable to the adversarial attack (``trashcan''$\rightarrow$``coffeepot''). The lower image has a more complete and accurate region of interest and thus retains the predicted label. (2) By examining the change of activation map from original to adversarial images, we found that a successful adversarial attack tends to deviate and scatter the attention area. The distraction of visual attention makes the prediction focus on incomplete and wrong regions.

These attention-oriented observations inspired us to design an adversarial defense solution by rectifying and 
preserving visual attention. The proposed Attention-based Adversarial Defense (AAD) framework consists of 
two components (as illustrated in Fig.~\ref{fig:4}): (1) attention rectification, to 
complete and correct the prediction of original images focusing on the actual object of interest; (2) attention 
preservation, to align the visual attention area between adversarial and original images with alleviating 
the feature divergence.
The main contributions of this study are three-fold:

\begin{itemize}
  \item We conducted a comprehensive data analysis and observed that a successful adversarial attack exploits the incomplete attention area and brings significant fluctuation to the activation map, which provides a new understanding of the adversarial attack problem from the attention perspective.
  \item A novel attention-based adversarial defense method is proposed to rectify and preserve the visual attention area simultaneously. Qualitative and quantitative results on MNIST, Cifar10, and ImageNet2012 demonstrate its superior defense performance. The framework is flexible and alternative modeling of attention loss can be readily integrated into the existing adversarial attack and defense solutions.
  \item In addition to applying visual attention to defense, we also discussed the possibility of using visual attention in adversarial attacks and gave a practical framework.
\end{itemize}

\section{Related Works}

Many approaches have been proposed to defend against adversarial samples~\cite{8884184, 8970483, Fangzhouy2017Defense,  madry2017towards, papernot2016distillation, guo2017countering, dhillon2018stochastic, song2017pixeldefend, tramer2017ensemble, 
liu2018feature, liao2017defense, shen2017ape, 8844865, 9206141, 9335499, 9286885, 8664462, 10.1145/3524619}. The attempts to develop adversarial defense solutions can be classified into three groups. (1) Denoising preprocessing, transforming the input samples before feeding them into the raw model, \emph{e.g.}, a generative adversarial 
network is proposed to eliminate the potential adversarial perturbation~\cite{shen2017ape}. This class of methods includes fascinating methods, such as using adversarial attacks as an image preprocessing method.~\cite{DBLP:conf/cvpr/XiaoZ20}
(2) Model modification, adding more layers or sub-networks and changing the loss functions of the raw model, \emph{e.g.}, Papernot \emph{et al.} designed a student network for knowledge distillation~\cite{hinton2015distilling} from the raw network and reduced the sensitivity to directional perturbations~\cite{papernot2016distillation}. 
(3) Adversarial training, adding adversarial samples into the training set to update the model parameters, \emph{e.g.}, Madry \emph{et al.} proposed to replace all clean images with adversarial images to protect against the adversary~\cite{madry2017towards}. Under the training framework of adversarial training, Adversarial Logits Pairing (ALP)~\cite{alp}, and TRADE~\cite{pmlr-v97-zhang19p} act more effectively. There is also a work to sort out recent benchmarking of adversarial training methods~\cite{DBLP:conf/cvpr/DongFYPSXZ20}. It is discussed in~\cite{athalye2018obfuscated} that the former two groups of defense methods work by obfuscating gradients, which provide only ``a false sense of security'' and have been successfully attacked by circumventing the gradient calculation. Adversarial training, although simple, does not rely on obfuscated gradients and has been proved to improve model robustness by correcting sample distribution~\cite{song2017pixeldefend}. 
At the same time, adversarial attacks have also shifted from the original single-step attack method and multi-step attack method to more automated hyperparameter finding methods, such as Square Attack~\cite{DBLP:conf/eccv/AndriushchenkoC20}, AutoAttack~\cite{DBLP:conf/icml/Croce020a}, MBbA~\cite{DBLP:conf/mm/DuanLX0X21} and MsGM~\cite{DBLP:journals/tomccap/DuanLDXT22}.
Previous defense work tends to use complex constraints and mathematical methods to constrain the model, while our work prefers to approach method design in a visually understandable way.

\section{Attention-oriented Data Analysis}\label{section2}
Visual attention helps explain to what extent each pixel of a given image contributes to the prediction of the network. Since the adversarial attack is designed to change the original prediction, we are motivated to examine the relationship between visual attention and adversarial attack. This data analysis section attempts to address the following two questions:

\begin{itemize}
  \item[1.]  What kinds of images are vulnerable to adversarial attacks?
  \item[2.]  How the visual attention of the adversarial image deviates from the original image?
\end{itemize}

Before presenting data analysis setting and observations, we first make agreements on several key terms:

\begin{itemize}
  \item \emph{Activation map}: In this study, we obtain the \emph{activation map} for a 
  given input image $x$ using Grad-CAM~\cite{selvaraju2017grad}~\footnote{To guarantee the derived data 
  observations are insensitive to the choice of activation map generator, we also employed LIME~\cite{ribeiro2016should} 
  for data analysis and obtained consistent observations.}, 
  which is denoted as:
  \begin{equation}
      g(x)=\mbox{Grad-CAM}(x).
  \end{equation}
    
  Grad-CAM is an interpretable method that uses the forward-propagated activation value $A$ of a specific layer of the model and the back-propagated gradient value $\alpha$. Its working process can be described as: 
  \begin{equation}
    L^c_{Grad-CAM}=\mbox{ReLU}(\sum_k \alpha^c_k A^k).
  \end{equation}
  where $k$ refers to the number of channels in the layer, and $c$ refers to the category displayed by Grad-CAM.
  
  Usually, the $L^c_{Grad-CAM}$ obtained by the Grad-CAM algorithm is a feature map smaller than the original image, which needs to be resized to draw the corresponding heat map. Some of the subsequent experiments in this paper need to use $L^c_{Grad-CAM}$ for IoU (Intersection-over-Union) comparison calculation. If there is no need to resize to the original image, we usually use $L^c_{Grad-CAM}$ to calculate directly.

  \item \emph{Attention area}:
      To prevent low-contribution pixels from affecting the analysis results, we introduce \emph{attention area} as 
      the binary mask indicating image pixels with contribution 
      value above a threshold $\kappa$~\footnote{We provide two methods for determining $\kappa$: (a) using a fixed value (b) using the mean of the attention map. In cases not explicitly stated in this paper, $\kappa$ uses the mean of the attention map.}:
      \begin{equation}
          Att(x)=g(x) \ge \kappa.
      \end{equation}
  \item \emph{Ground-truth area}: Taking the classification task as an example, the attention area corresponds to the region that relies on recognizing a specific object by the deep learning model. This study uses \emph{ground-truth area} to indicate the object region obtained by the object bound-box. The ground-truth area of object $l$ in image $x$ is denoted as $GT_l(x)$.
  \item \emph{Adversarial attack}: The adversarial samples used in data analysis, unless otherwise specified, are 
  generated by $StepLL$~\cite{tramer2017ensemble}:
      \begin{equation}\label{eq:4}
          x_{adv}=x_{ori}-\varepsilon \cdot sign(\bigtriangledown_xL(f(x_{ori}), y_{LL}))
      \end{equation}
      where $x_{ori}, x_{adv}$ represent original and adversarial images, $\varepsilon$ is the step size, $f(\cdot)$ is 
      the original network, and $y_{LL}$ denotes the label with the lowest confidence.
  \end{itemize}

  \begin{table}[t]
    \begin{center}
    \begin{tabular}{|c|c|c|c|}
    \hline
    \multicolumn{2}{|c}{$\mathcal{I}_{succeed}$} & \multicolumn{2}{|c|}{$\mathcal{I}_{fail}$} \\
    \hline
    Percentage & IoU$_{Att\_GT}$ & Percentage & IoU$_{Att\_GT}$ \\
    \hline
    74.1\% & 0.647 & 25.9\% & 0.701 \\
    \hline
    \end{tabular}
    \end{center}
    \caption{Average IoU between attention and ground-truth area.}
    \label{tab:analysis1}
\end{table}

\subsection{Adversarial Attack Vulnerability}\label{analysis1}
It is noticed that adversarial attack not always succeeds and fails on some samples.
This motivates us to study what 
characteristics make these samples robust to the attack and retain the original decision. Specifically, we examined 
the attention area of different images in the context of the classification problem and analyzed its correlation with 
the vulnerability to adversarial attack.
The data analysis was conducted with the classification network, InceptionV3~\cite{szegedy2016rethinking}, over 
the 50,000 images in the development set of ImageNet 2012~\cite{deng2009imagenet}. Since we view visual attention 
as support on the most confident output, the 38,245 development images with the correct top-1 prediction construct the 
image set $\mathcal{I}_{att}$ for attention analysis.

For each image, $x\in\mathcal{I}_{att}$, its activation map $g(x)$, and the ground-truth area $GT_l(x)$ corresponding to the correct label $l$ was extracted. To examine whether the visual attention matches the actual object region, we compared the attention and ground-truth areas. The attention area was extracted by selecting image-specific threshold $\kappa$, which makes $Att(x)$ and $GT_l(x)$ have 
the same area size. The IoU between attention and ground-truth area was calculated as follows:
\begin{equation}
  IoU_{Att\_GT}(x)=\frac{Att(x)\bigcap GT_l(x)}{Att(x)\bigcup GT_l(x)}
\end{equation}

We separate images from $\mathcal{I}_{att}$ into two subsets, those retaining the original decision where the adversarial attack failed to construct $\mathcal{I}_{fail}$, and those changing decision where adversarial attack succeeded to construct $\mathcal{I}_{succeed}$. The percentage of images falling in each subset and the corresponding average $IoU_{Att\_GT}$ are summarized in Table~\ref{tab:analysis1}. Since the original network correctly classifies all the images, both subsets show large IoU scores. Between the two subsets, $\mathcal{I}_{fail}$ obtains higher IoU than $\mathcal{I}_{succeed}$. Focusing on the 5,857 images with IoU$<0.5$, we examined the percentage of images falling in each subset and the average confidence score on the correct label of the original images. The results are reported in Table~\ref{tab:analysis1-2}. When IoU is less than 0.5, the model does not focus on the foreground area of the target and does not use task-related features. However, at this time, the model still maintains high confidence, which shows that the model uses many image features that are not relevant to the task. Combining results from Table~\ref{tab:analysis1} and Table~\ref{tab:analysis1-2}, we observed that the images with low IoU tend to obtain high confidence scores based on the task-unrelated features and have higher vulnerability to be adversarially attacked.

\begin{table}[t]
  \begin{center}
  \begin{tabular}{|c|c|c|c|}
  \hline
  \multicolumn{2}{|c}{$\mathcal{I}_{succeed}$} & \multicolumn{2}{|c|}{$\mathcal{I}_{fail}$} \\
  \hline
  Percentage & Top 1 Confidence & Percentage & Top 1 Confidence \\
  \hline
  82.8\% & 0.8944 & 17.2\% & 0.7941 \\
  \hline
  \end{tabular}
  \end{center}
  \caption{Percentage and average confidence score of images with IoU$<0.5$.}\label{tab:analysis1-2}
\end{table}

\subsection{Attention Deviation from Adversarial Attack}\label{analysis2}
Adversarial samples only impose small perturbations on the original input but encounter significant changes in the output prediction. This motivates us to explore what factors contribute to the non-trivial output change. This subsection studies the attention deviation due to adversarial attacks and examines the consistency of the attention area between original samples and adversarial samples.

We utilized the same image set $\mathcal{I}_{att}$ for data analysis. Assuming $x_{adv}$ represents the adversarial 
sample generated by attacking original sample $x_{ori}$, $Att(x_{ori}),Att(x_{adv})$ respectively denote the 
attention area of original and adversarial samples. The raw activation map generated by Grad-CAM constitutes 
an $8\times 8$ grid. The attention area of original and adversarial samples is constructed by keeping the same 
number of grid cells with the highest attention score. Under a certain number of grid cells $\#cell$, the IoU 
of attention area between original samples and adversarial samples was calculated as follows:

\begin{equation}
    IoU_{ori\_adv}(x)=\frac{Att(x_{adv})\bigcap Att(x_{ori})}{Att(x_{adv})\bigcup Att(x_{ori})}
\end{equation}

\begin{figure}[t]
  \begin{center}
     \includegraphics[width=0.7\linewidth]{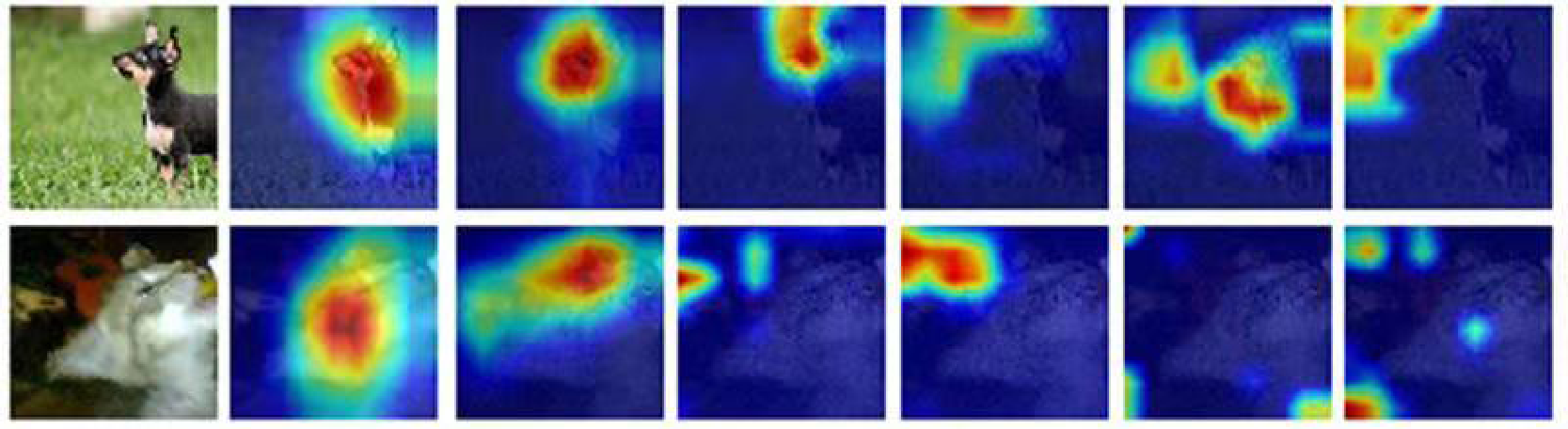}
  \end{center}
     \caption{The change of activation map in iterative attack. (From left to right columns: original image, 
     activation map for the original image and adversarial images after 1--5 rounds of attack.)}
  \label{fig:2}
\end{figure}

Varying the number of remaining grid cells from 5 to 30, we summarized the average $IoU_{ori\_adv}$ in Table~\ref{tab:analysis2} for $\mathcal{I}_{succeed}$ and $\mathcal{I}_{fail}$ respectively. We find a 
consistent result for different selections of grid cells: the IoU score of failed attack group is significantly higher than that of the successful attack group. The heavy attention deviation of adversarial samples from original samples offers a strong indication of the successful attack. Other than the decrease of overlap of attention areas, it is also evidenced from Fig.~\ref{fig:1},\ref{fig:2} and other samples that a successful adversarial attack tends to make attention scattered. A possible explanation for these observations is that successful adversarial perturbation on the input misleads the output prediction by distracting and scattering the original attention.

\begin{table}[t]
  \begin{center}
  \begin{tabular}{|c|c|c|}
  \hline
   \#cell & IoU$_{ori\_adv}$ for $\mathcal{I}_{succeed}$ & IoU$_{ori\_adv}$ for $\mathcal{I}_{fail}$ \\
  \hline
   5 & 0.393 & 0.574 \\ \hline
   10 & 0.490 & 0.677 \\ \hline
   20 & 0.609 & 0.766 \\ \hline
   30 & 0.676 & 0.807 \\
  \hline
  \end{tabular}
  \end{center}
  \caption{Average attention IoU between original and adversarial samples.}\label{tab:analysis2}
\end{table}

\subsection{LIME-based Attention Data Analysis}
We conducted attention-oriented data 
analysis by using Grad-CAM~\cite{selvaraju2017grad}. To guarantee 
the derived data observations are insensitive to the choice of 
activation map generator, we also used another attention generation 
method for attention data analysis, and this section introduces the 
data analysis results.

\begin{figure}[!th]
  \begin{center}
     \includegraphics[width=0.5\linewidth]{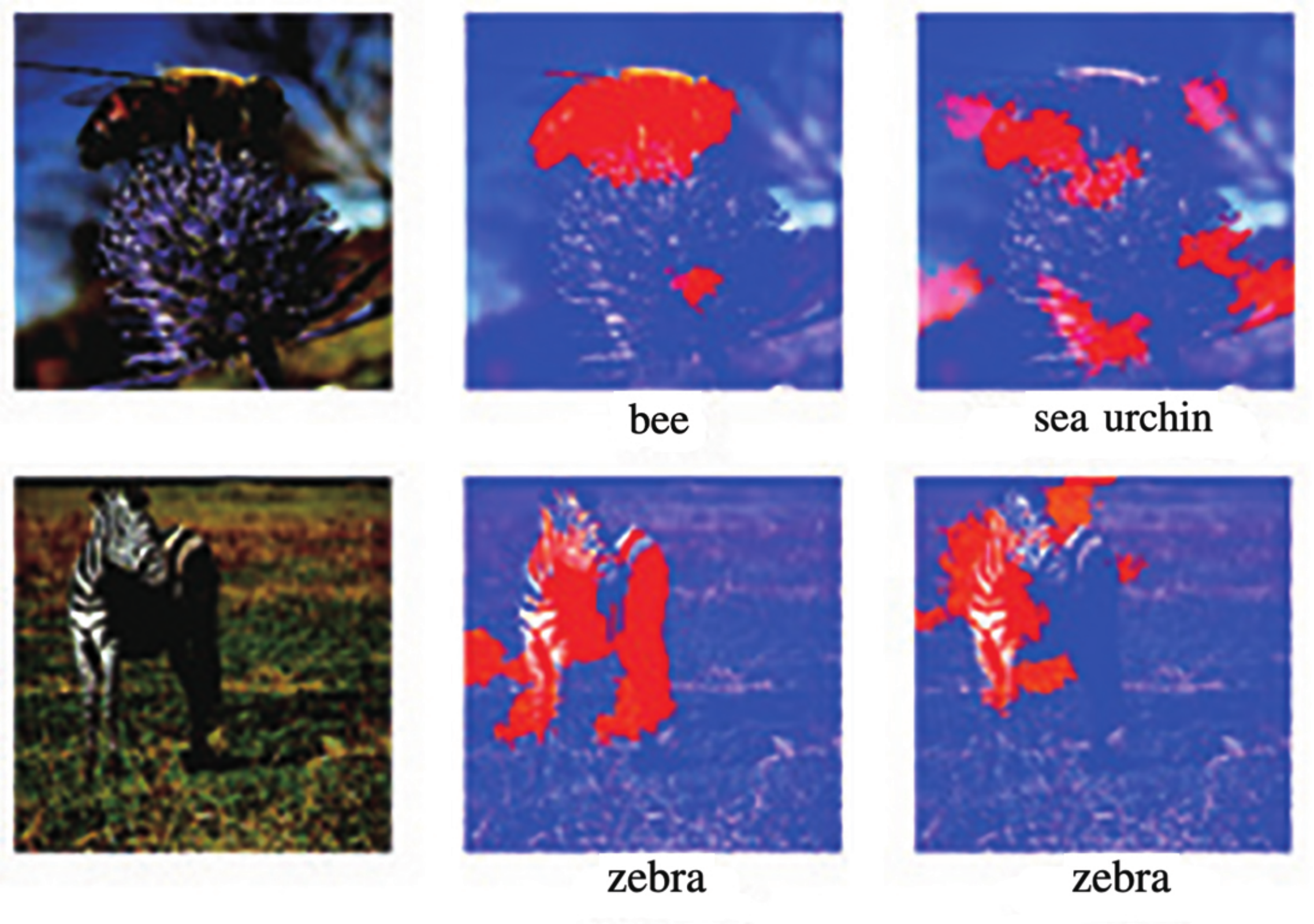}
  \end{center}
     \caption{Visual attention of successful-attacked image (top row) and failed-attacked image (bottom row) by LIME.  In each row we show the original image, original image's activation map and adversarial image's activation map.}
  \label{fig:a1}
\end{figure}

LIME learns an interpretable model locally around the prediction, 
to explain the predictions of classifiers in an interpretable and 
faithful manner~\cite{ribeiro2016should}. Similar to Sec.~\ref{analysis1} and 
Sec.~\ref{analysis2}, we use LIME to re-examine the observations 
concerning adversarial attack vulnerability and attention deviation 
from adversarial attack. Fig.~\ref{fig:a1} illustrates the LIME-generated 
activation map by retaining the top 8 features. For the successfully 
attacked images (top row), a similar phenomenon of shrinking 
and scattered attention area is observed.

By analyzing the same 38,245 images from $\mathcal{I}_{att}$, we summarize the respective statistics regarding average IoU in Table~\ref{tab:1} and Table~\ref{tab:2}. Consistent observations are obtained with a different number of retained features: (1) images vulnerable to adversarial attack tend to have a lower IoU score; (2) successful adversarial attack deviates the activation map of adversarial images from original images.

\begin{table}[h]
  \begin{center}
  \begin{tabular}{|c|c|c|}
  \hline
  \#.Features & $\mathcal{I}_{succeed}$ & $\mathcal{I}_{fail}$ \\
  \hline
   8 & 0.413 & 0.442 \\ \hline
   10 & 0.484 & 0.516 \\
  \hline
  \end{tabular}
  \end{center}
  \caption{Average attention IoU between attention and ground-truth area.}\label{tab:1}
\end{table}

\begin{table}[h]
  \begin{center}
  \begin{tabular}{|c|c|c|}
  \hline
  \#.Features & $\mathcal{I}_{succeed}$ & $\mathcal{I}_{fail}$ \\
  \hline
   8 & 0.138 & 0.212 \\ \hline
  10 & 0.159 & 0.241 \\
  \hline
  \end{tabular}
  \end{center}
  \caption{Average attention IoU between adversarial images and corresponding original images.}\label{tab:2}
\end{table}

\section{Attention-based Adversarial Defense}

The above data analysis demonstrates that successful adversarial perturbation leads to significant visual 
attention change. Our defense solution is therefore motivated to restrict the attention change to improve 
adversarial robustness. Specifically, observations from data analysis correspondingly inspire the two components 
in the proposed attention-based adversarial defense framework (as illustrated in Fig.~\ref{fig:4}): 
(1) attention rectification, to guide the attention area of original samples to the area where the real object exists; 
(2) attention preservation, to punish the deviation of attention area from adversarial samples to original samples.

\subsection{Attention Rectification}
As evidenced from Table~\ref{tab:analysis1}, it is more vulnerable to adversarial attacks for those samples whose prediction relies on the unrelated region instead of the ground-truth area. One possible explanation is that these samples failing to focus on the actual region of interest suffer more from the adversarial perturbations and have a higher risk of being misclassified. Therefore, our first component motivates the model to focus more on the ground-truth area for prediction.

Since the ground-truth area is generally unavailable during the training process, we turn to rectify the completeness of the attention area. The idea is that the attention area should include all the regions critical for prediction. In other words, the regions beyond the attention area are expected to contribute trivially to the correct prediction.

\begin{figure*}[t]
  \begin{center}
     \includegraphics[width=1.0\linewidth]{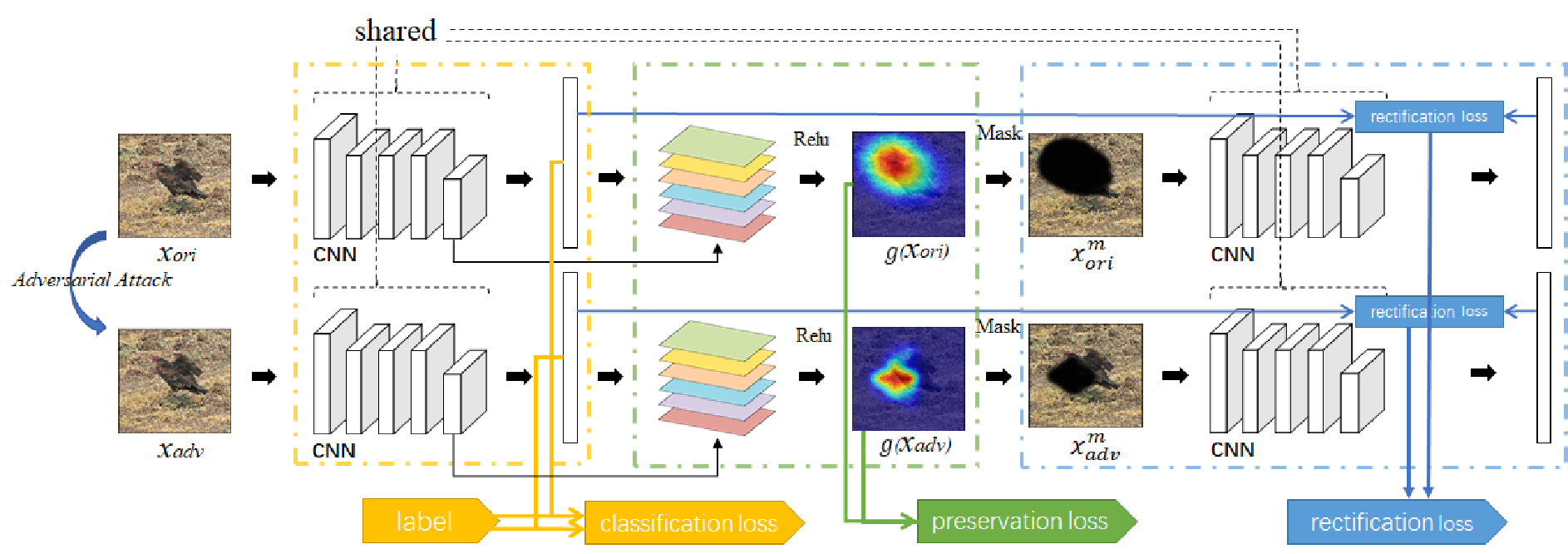}
  \end{center}
     \caption{The proposed attention-based adversarial defense framework. The upper and lower parts correspond to the training over original and adversarial samples. The change of the activation map is simultaneously constrained by the rectification loss, preservation loss, and classification loss. The four convolutional neural networks share the same parameters.}
  \label{fig:4}
\end{figure*}

To realize this, we integrate the generation of attention area into the end-to-end training process. As illustrated 
in the upper part of Fig.~\ref{fig:4}, a hard mask is imposed according to the extracted attention area:
\begin{equation}
    x^m=x\odot[1-Att(x)],
\end{equation}
where $x$ is the original image, $x^{m}$ denotes the image after the mask, and $\odot$ denotes element-wise multiplication. 
To guarantee all critical regions are excluded from $x^m$, it is desired that $x$ and $x^{m}$ lead to the prediction results 
as different as possible. Therefore, by feeding $x$ and $x^{m}$ into the same convolutional network $f(\cdot)$ to 
obtain the prediction vector $f(x)$ and $f(x^m)$, we expect the difference between $f(x)$ and $f(x^m)$ to be as large 
as possible. The same constraint is added to the adversarial image. For each original image $x$ and the corresponding 
adversarial image $x_{adv}$, the goal is to minimize the following rectification loss:
\begin{equation}\label{eq:3}
    L_{r}(x)=-\big [\mathcal{L}(f(x),f(x^{m}))+\mathcal{L}(f(x_{adv}),f(x^{m}_{adv}))\big ]
\end{equation}
where $\mathcal{L}(\cdot,\cdot)$ denotes a certain distance measure between two vectors.

\subsection{Attention Preservation}
It is observed from Sec.~\ref{analysis2} that the shifted prediction results of the adversarial image partially owe to the deviation of attention from the original image. This component attempts to preserve the activation 
map between original and adversarial images to reduce the influence of input perturbation on the prediction result.

The original image $x$ and adversarial image $x_{adv}$ are issued to the same convolutional network to obtain activation map $g(x)$ and $g(x_{adv})$. A preservation loss is designed to minimize the pairwise difference between the two activation maps:
\begin{equation}\label{eq:2}
    L_{p}(x)=\mathcal{L}\big ( g(x),g(x_{adv})\big )
\end{equation}

Combining with the rectification loss defined in the previous subsection, the overall loss for the given original 
image $x$ is calculated as follows:
\begin{equation}\label{eq:1}
        L_{total}(x) = L_{c}(x) + \alpha L_{r}(x) + \beta L_{p}(x),
\end{equation}
where $L_{c}$ is for correct classification and defined as the standard multi-label soft margin 
loss, $\alpha$ and $\beta$ are weight parameters reflecting the importance of the respective component.

The three losses in Eqn.~\eqref{eq:1} jointly regularize the original and adversarial image's attention to the critical regions for prediction. This provides a general attention-based framework to improve adversarial robustness by rectifying and preserving visual attention. Alternative realizations of rectification and preservation loss are compatible with the framework.

\begin{figure*}[!tb]
  \begin{center}
    \includegraphics[width=1.0\linewidth]{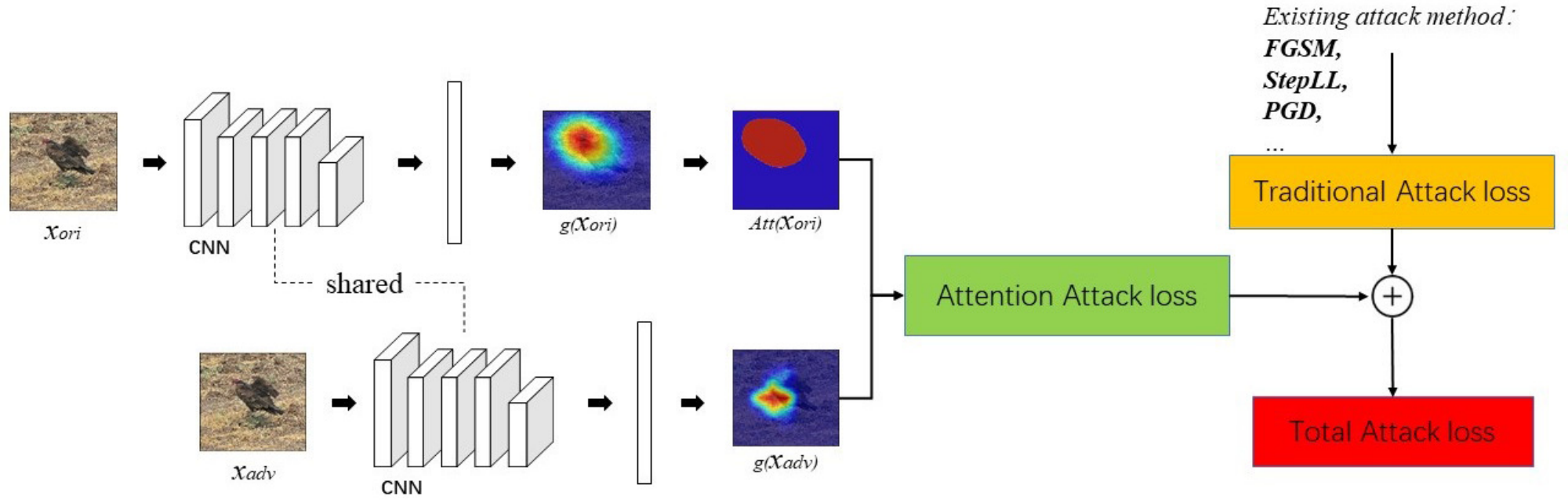}
    \center
    \caption{The framework of A$^3$Attack.}
    \label{fig:11}
  \end{center}
\end{figure*}

\subsection{Attention-Augmented Adversarial Attack} \label{AAA}

Data analysis showed that attention deviation contributes significantly to a successful adversarial attack. The proposed AAD method exploits this observation to improve the robustness of potential adversarial attacks. From a counter perspective, this observation also inspires the design of new attack methods by considering attention.

We introduce a general Attention-Augmented Adversarial Attack (A$^3$Attack) framework to be compatible with existing adversarial attack methods using gradient information. The overall framework is shown in Fig.~\ref{fig:11}. When conducting an $A^3$ attack, the original image needs to be sent into the neural network to calculate the attention area. In order to better calculate the distance, we have binarized the attention map. Then, the image is sent to the neural network to obtain the attention map of the image. At this time, we separate the two attention maps as a loss and add them to the adversarial loss to update the perturbation of the adversarial examples to obtain more robust adversarial examples. The attention area of the original and adversarial samples are extracted, and the difference between the two attention areas is utilized to define attention attack loss and encourage attention derivation in the adversarial samples to be generated. The total attack loss combines the new attention attack loss and the traditional attack loss from existing attack methods, \emph{e.g.}, FGSM, Stepll, PGD, \emph{etc.}.

$l_1$-norm can be used to calculate the attention difference, and the attention attack loss is defined as:
\begin{equation}
    Attention Attack Loss=- \Vert Att(x)-g(x)\Vert_{1}
\end{equation}
Taking StepLL as an example of existing attack methods, the total attack loss 
can be represented as:
\begin{equation}
    TotalAttack Loss=L(f(x),y_{LL})-\sigma \Vert Att(x)-g(x)\Vert_{1}
    \end{equation}
where {$\sigma$} is the weight parameter controlling the contribution of attention deviation. The adversarial sample of StepLL+A$^3$Attack is generated as follows:
\begin{equation}
    x_{adv}=x_{ori}-\varepsilon  \cdot sign[\triangledown_{x}(TotalAttack Loss)]
    \end{equation}

In this way, in addition to changing the prediction score of the most confident class, the adversarial 
perturbation is also designed to deviate the attention area.

\section{Experiment}

\subsection{Experimental Setting}
The proposed Attention-based Adversarial Defense (AAD) framework can be 
applied to different convolutional networks. In this study, we conducted 
qualitative and quantitative experiments with three networks, LeNet on 
MNIST dataset, ResNet9~\cite{resnet} on Cifar10 dataset~\cite{krizhevsky2009learning} and
ResNet50~\cite{resnet} on ImageNet2012~\cite{deng2009imagenet}. For LeNet, the primary parameters are empirically 
set as follows: batchsize = 100, learning rate = 0.001, epoch = 5, keeprate = 0.5, 
$\kappa = mean(g(x))$. For ResNet9, we employ the commonly used parameters 
as: batchsize = 100, learnning rate = 0.0001, epoch = 200, keeprate = 0.5, 
$\kappa = mean(g(x))$. For ResNet50, we employ the commonly used parameters 
as: minibatchsize = 128, using pre-trained model~\footnote{For the fairness of the experimental comparison, we only use the pre-trained model trained on ImageNet2012.} and set the learning rate of 0.045, decayed by a factor of 0.94 every two epochs.
For distance measure $\mathcal{L}(\cdot,\cdot)$ in attention losses calculation, 
we used $l_2$-norm in Eqn.\eqref{eq:3} to increase the distance between the soft labels' output by the model for $x$ and $x_m$ 
and $l_1$-norm in Eqn.\eqref{eq:2} to encourage sparsity~\footnote{Alternative distance 
measures are also allowed to encourage different characteristics of the activation map.}.
For the generation of adversarial samples, we use the same settings as Madry \emph{et al.}~\cite{madry2017towards} when using PGD.

\subsection{Visualization of Activation Map Evolution}
\begin{figure}[t]
  \begin{center}
     \includegraphics[width=0.7\linewidth]{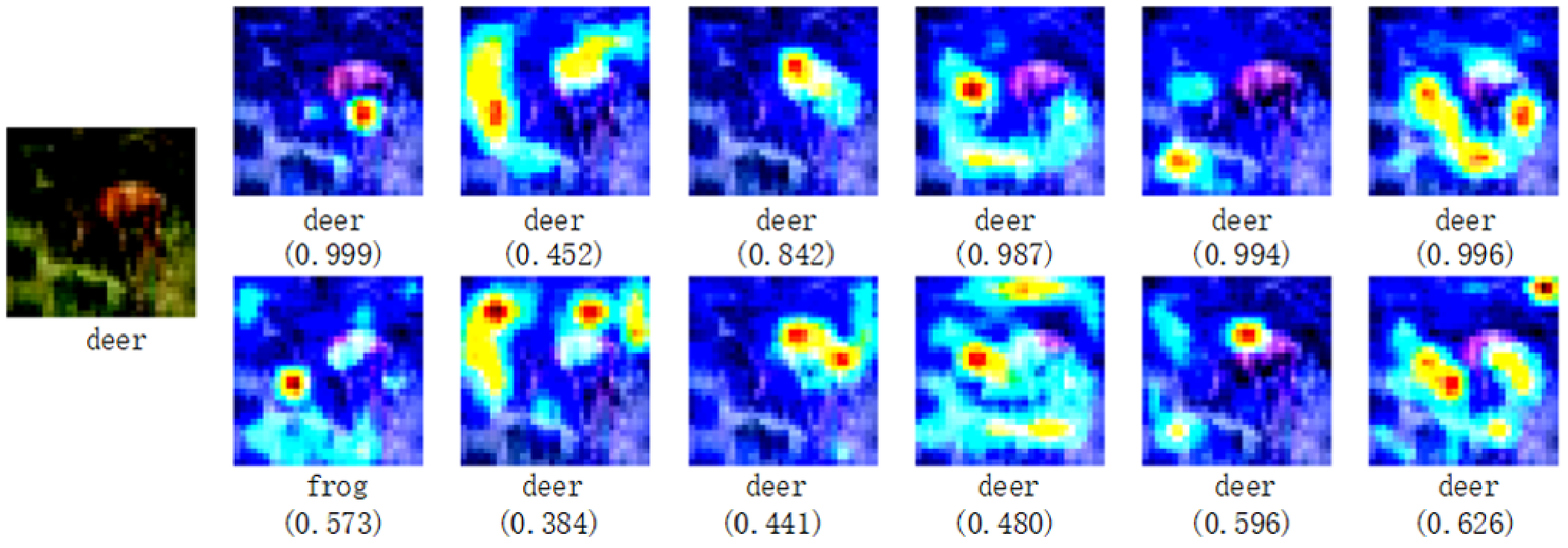}
     \includegraphics[width=0.7\linewidth]{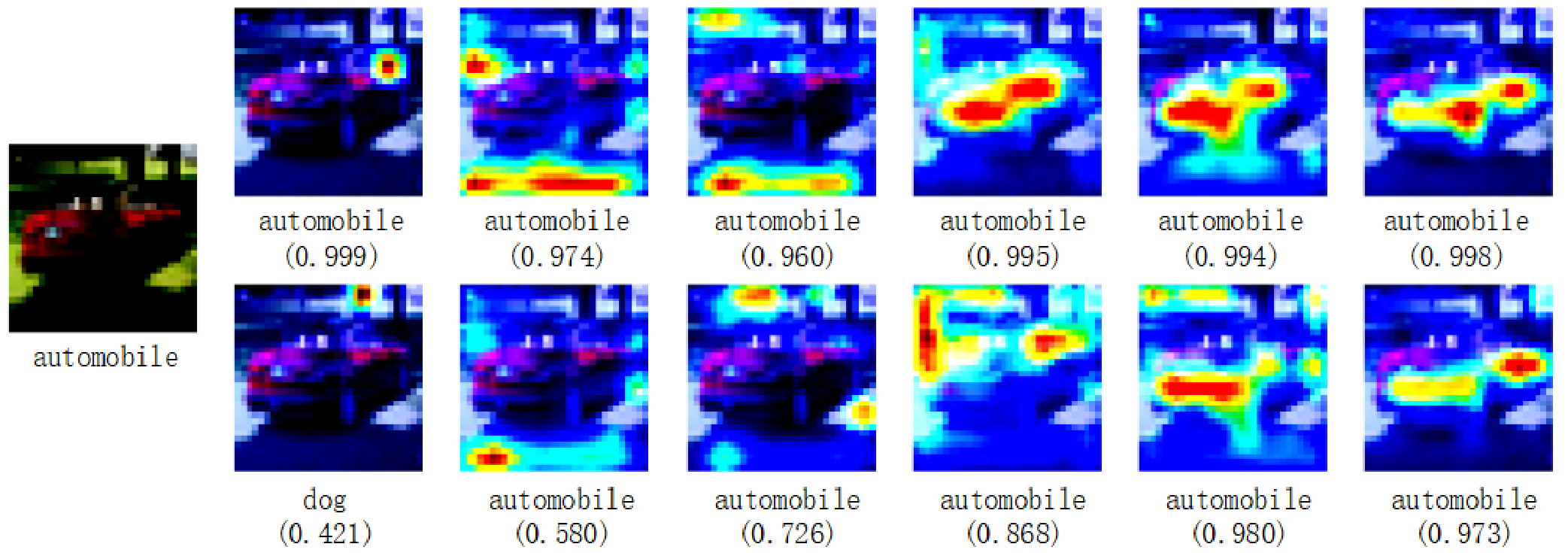}
  \end{center}
     \caption{Activation map evolution in different training epochs. For each example image on the left, we show on the right the activation maps of its original image (top row) and adversarial image (bottom row). The six activation maps are extracted at training epochs $0, 10000, 20000, 30000, 40000$, and $50000$. Below each activation map shows the predicted object label and the corresponding confidence score.}
  \label{fig:5}
\end{figure}

The proposed AAD framework is expected to adapt the network to make predictions of both original and adversarial images focusing on the critical regions. Fig.~\ref{fig:5} visualizes the evolution of the activation map for two Cifar10 images as the training epoch increases.

It is shown that for each example image on the top row, the attention area of the original images is gradually rectified to the object of interest. The prediction confidence first reduces due to attention rectification loss and preservation loss and then recovers to guarantee high prediction confidence and complete attention area. On the second row, the raw attention area of the adversarial image deviates much from the original image (first column on the right), which leads to the misclassification at the beginning (``deer''$\rightarrow$``frog'', ``automobile''$\rightarrow$``dog''). As the adversarial training proceeds, the attention area of the adversarial image becomes consistent with that of the original image and together fits onto the object of interest at last. The confidence score for the correct object class also increases as the attention area evolves to demonstrate the improvement of robustness against potential attacks.

\begin{figure}[t]
\begin{minipage}[c]{0.4\linewidth}
\centering
\includegraphics[width=1.0\textwidth]{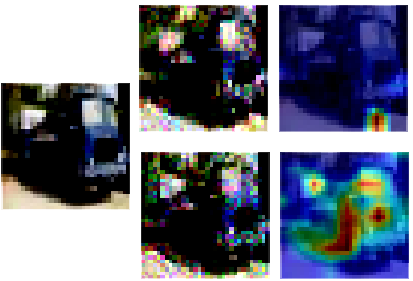}
\centerline{(a)}
\end{minipage}
\begin{minipage}[c]{0.4\linewidth}
\centering
\includegraphics[width=1.0\textwidth]{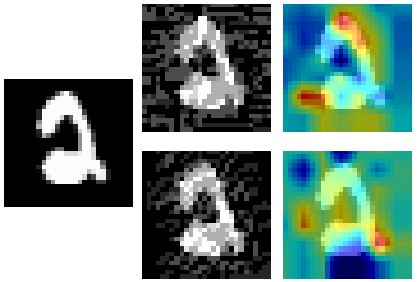}
\centerline{(b)}
\end{minipage}
\caption{Example images from Cifar10 (a) and MNIST (b). The left shows the original image. The top row on the right shows the adversarial image generated from PGD with a standard adversarial training method and its corresponding activation map. The bottom row shows the results of adversarial training with the proposed AAD method.}
 \label{fig:6}
 \end{figure}

\subsection{Performance on Adversarial Defense}

\begin{table*}[!h]
  \begin{center}
    \resizebox{\textwidth}{!}{
    \begin{tabular}{|c|c|c|c|c|c|c|c|c|c|c|}
    \hline
    &\multicolumn{10}{|c|}{MNIST} \\ \cline{2-11}
    & Clean & FGSM & PGD10 & PGD20 & PGD40 & PGD100 & CW10 & CW20 & CW40 & CW100 \\ \hline
    RAW & 99.18\% & 7.72\% & 0.49\% & 0.54\% & 0.64\% & 0.53\% & 0.0\% & 0.0\% & 0.0\% & 0.0\% \\ \hline
    Madry adv &	99.00\% & 96.92\%	& 95.69\% & 95.52\% & 94.14\% & 92.49\% & 95.69\% & 92.69\% & 87.67\% & 85.17\% \\ \hline
    Madry adv + AAD & \textbf{99.20\%} & 97.12\% & 98.02\% & 96.81\% & 94.77\% & 93.02\% & \textbf{97.28\%} & 95.04\% & 91.31\% & 89.77\% \\ \hline
    Kannan ALP & 98.61\% & 97.59\% & 98.23\% & 97.52\% & 96.04\% & 94.68\% & 96.96\% & 95.48\% & 93.18\% & 92.39\% \\ \hline
    Kannan ALP + AAD & 98.52\% & \textbf{97.60\%} & \textbf{98.24\%} & \textbf{97.56\%} & \textbf{96.11\%} & \textbf{94.82\%} & 97.18\% & \textbf{95.62\%} & \textbf{93.54\%} & \textbf{92.72\%} \\ \hline    
  \end{tabular}
    }
    \end{center}

    \begin{center}
      \resizebox{\textwidth}{!}{
      \begin{tabular}{|c|c|c|c|c|c|c|c|c|c|c|}
      \hline
      &\multicolumn{10}{|c|}{Cifar10} \\ \cline{2-11}
      & Clean & FGSM & PGD10 & PGD20 & PGD40 & PGD100 & CW10 & CW20 & CW40 & CW100 \\ \hline
      RAW & \textbf{90.51\%} & 14.24\% & 5.93\% & 5.80\% & 5.97\% & 6.04\% & 0.0\% & 0.0\% & 0.0\% & 0.0\% \\ \hline
      Madry adv & 83.36\% & 63.28\%	&  49.29\% & 45.25\% & 44.85\% & 44.72\% & 37.21\% & 36.27\% & 35.92\% & 35.79\% \\ \hline
      Madry adv + AAD & 83.62\% & 64.98\% & 50.13\% & 48.99\% & 48.55\% & 48.49\% & 40.83\% & 39.90\% & 39.75\% & 39.64\% \\ \hline
      Kannan ALP & 82.80\% & 64.96\% &52.79\% & 49.04\% & 48.81\% & 48.60\% & 41.35\% & 40.43\% & 40.11\% & 40.00\% \\ \hline
      Kannan ALP + AAD & 80.50\% & \textbf{65.97\%} & \textbf{53.43\%} & \textbf{52.34\%} & \textbf{52.14\%} & \textbf{51.99\%} & \textbf{43.32\%} & \textbf{42.60\%} & \textbf{42.43\%} & \textbf{42.34\%} \\ \hline  
    \end{tabular}
      }
      \end{center}
      
      \begin{center}
        \resizebox{\textwidth}{!}{
        \begin{tabular}{|c|c|c|c|c|c|c|c|c|c|c|}
        \hline
        &\multicolumn{10}{|c|}{ImageNet-2012} \\ \cline{2-11}
        & Clean & FGSM & PGD10 & PGD20 & PGD40 & PGD100 & CW10 & CW20 & CW40 & CW100 \\ \hline
        RAW & \textbf{75.60\%} & 4.88\% & 0.50\% & 0.43\% & 0.42\% & 0.33\% & 3.28\% & 2.50\% & 1.93\% & 1.65\% \\ \hline
        Madry adv & 55.37\% & 9.49\% & 2.88\% & 0.46\% & 0.05\% & 0.0\% & 7.58\% & 3.58\% & 1.38\% & 0.28\% \\ \hline
        Madry adv + AAD & 46.48\% & 11.47\% & 10.11\% & 5.29\% & 2.65\% & 1.3\% & 21.22\% & 19.73\% & 18.75\% & 18.05\% \\ \hline
        Kannan ALP & 50.68\% & 9.94\% & 23.67\% & 19.15\% & 14.89\% & 9.94\% & 26.15\% & 11.22\% & 6.3\% & 3.93\% \\ \hline
        Kannan ALP + AAD & 47.06\% & \textbf{14.86\%} & \textbf{29.15\%} & \textbf{22.45\%} & \textbf{17.56\%} & \textbf{12.59\%} & \textbf{37.40\%} & \textbf{34.46\%} & \textbf{32.90\%} & \textbf{31.37\%} \\ \hline   
      \end{tabular}
        }
        \end{center}
 \caption{Defense performance comparison with Madry et al.~\cite{madry2017towards} and Kannan \emph{et al.}~\cite{alp} under different 
 adversarial attack methods.}\label{tab:3}
\end{table*}

To demonstrate the defense performance of the proposed AAD framework, we compare its classification accuracy with a clean model and the defense model from Madry \emph{et al.}~\cite{madry2017towards} and adversarial logits pairing~\cite{alp}. Three additional attack methods are also employed to generate adversarial samples: FGSM, PGD($\epsilon = 0.3$ in MNIST, $\epsilon = 8pix$ in Cifar10 and $\epsilon = 16pix$ in ImageNet2012) and CW($\epsilon = 0.3$ in MNIST, $\epsilon = 8pix$ in Cifar10 and $\epsilon = 16pix$ in ImageNet2012)~\cite{kurakin2016adversarial,KurakinGB17,nicholas2017towards}. 

On the MNIST dataset, the Madry+AAD algorithm performs 1\% better than Madry's method under multiple attacks. ALP+AAD achieves slightly better performance than the ALP method. It is perceived that the handwritten digit in MNIST is classified as positioned in the center and covers a dominant image region. Fig.~\ref{fig:6}(b) shows such an example. In this case, the benefit of rectifying and preserving attention is limited. The proposed attention-based defense framework is more suitable for images with arbitrary object sizes and complex backgrounds. 

On the Cifar10 dataset, the proposed AAD method achieves superior defense performance under all the attack methods. The AAD algorithm plays a perfect role in defense of the CW algorithm. Fig.~\ref{fig:6}(a) shows an example image with its adversarial image and activation map. It is observed that standard adversarial training relies on the attention region beyond the object of interest for prediction. AAD manages to rectify and preserve the attention even for the strong adversarial image. Moreover, considering multi-round training adversarial samples further improves AAD's robustness to iterative attacks.

On the ImageNet2012 dataset, although AAD has a slight loss of precision on the clean sample, it has stronger robustness, especially in the performance of iterative attacks. Comparing the ALP results with the ALP+AAD results, it can be seen that the model with AAD has greatly improved the defense effect of CW and multiple iterations of PGD attacks.

We can observe that no matter which kind of adversarial training, the defense performance is obviously improved after using the AAD framework.

\begin{figure*}[!tb]
  \begin{minipage}[b]{0.48\linewidth}
  \centering
  \includegraphics[width=0.8\textwidth]{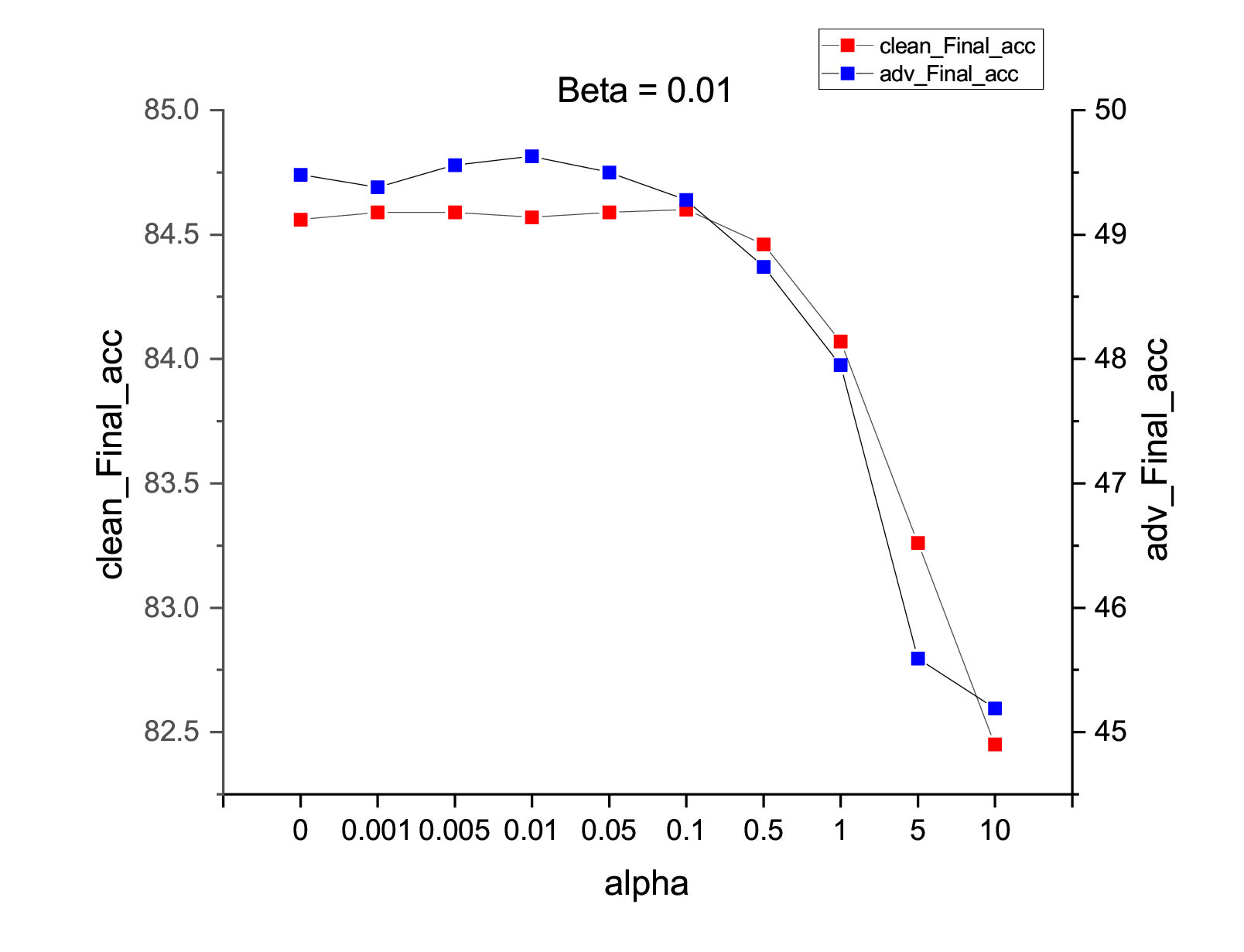}
  \centerline{(a) rectification loss}
  \end{minipage}
  \begin{minipage}[b]{0.48\linewidth}
  \centering
  \includegraphics[width=0.8\textwidth]{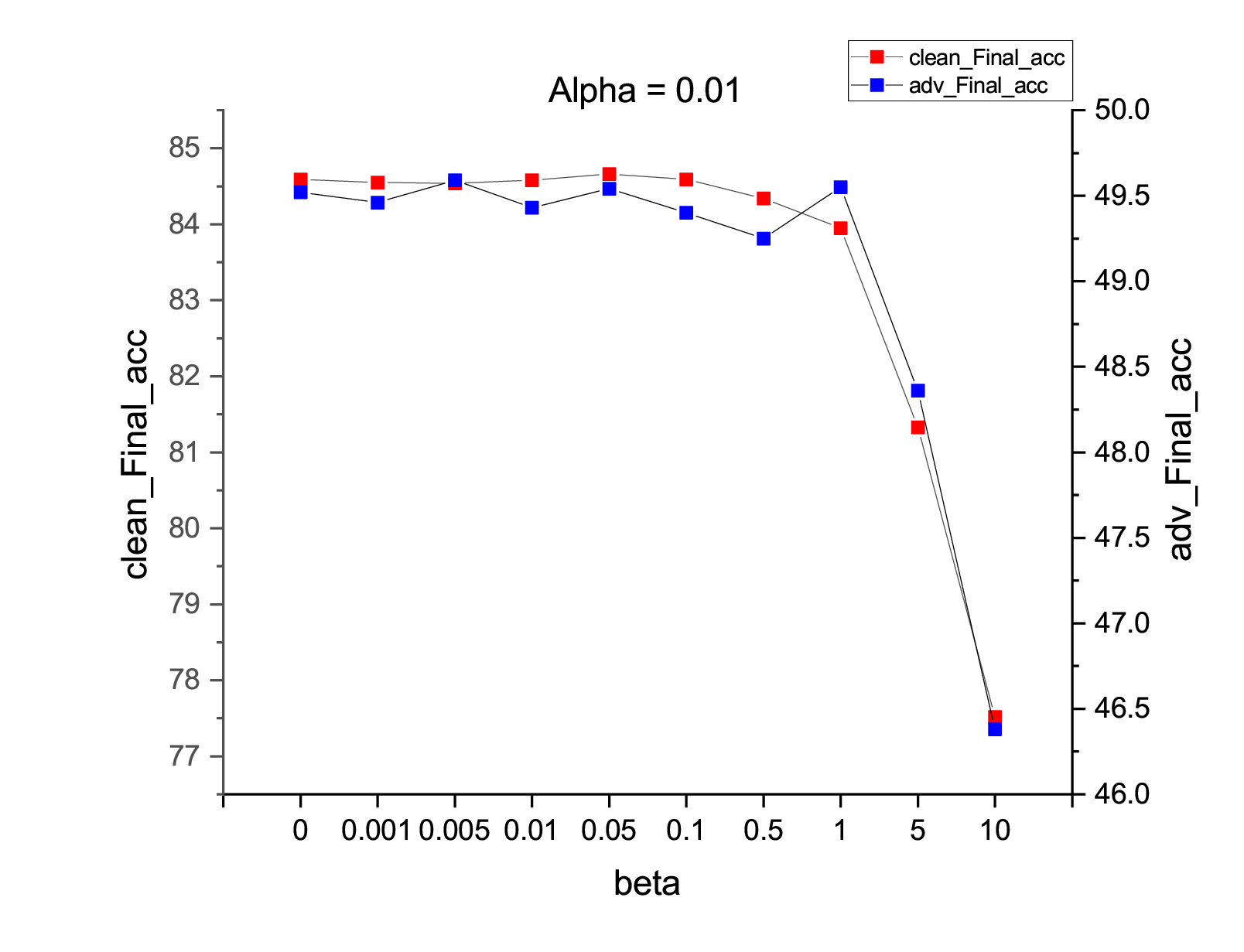}
  \centerline{(b) preservation loss}
  \end{minipage}
  \caption{Adversarial defense performance with different weight parameter configurations.
  The axis on the left represents the classification accuracy, and the one on the right represents the accuracy of adversarial samples.}
   \label{fig:7}
\end{figure*}

\begin{figure}[!t]
  \begin{center}
     \includegraphics[width=0.6\linewidth]{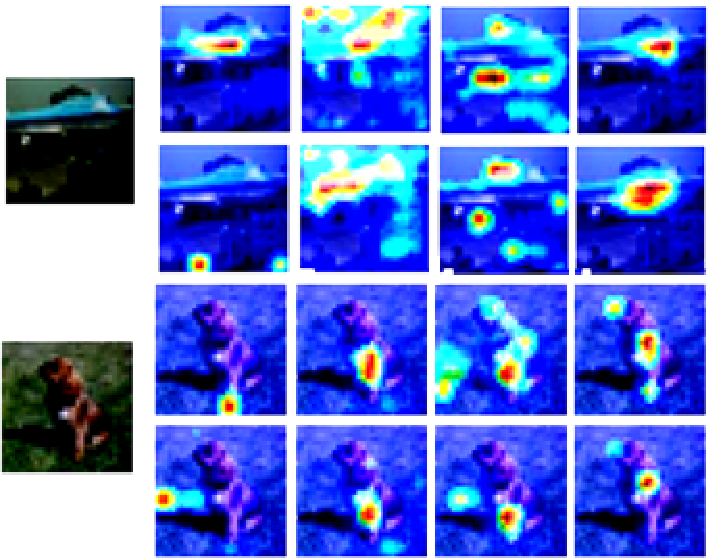}
  \end{center}
     \caption{Activation maps from the raw model, model considering rectification loss, model considering preservation 
     loss, and model with both rectification and preservation losses. }
  \label{fig:8}
 \end{figure}
  
\subsection{Parameter Sensitivity Analysis}

The proposed attention-based defense framework mainly involves the weight parameter $\alpha$ and $\beta$ in Eqn.~\eqref{eq:1}. This subsection serves to analyze the performance sensitivity to these parameters.
  
We first adjusted the weight parameter to analyze the contribution of the respective loss. The weight parameter sensitivity analysis experiment is conducted by fixing one of the weights and tuning the other weight. In Fig.~\ref{fig:7}(a)--(b), we show the defense classification accuracy of Cifar10 against different attack methods by tuning $\alpha$ and $\beta$, respectively.
  
When setting the weight for respective loss as $0$, \emph{i.e.}, excluding the corresponding rectification and preservation constraint, the classification accuracy curves experience a consistently significant decline when $\alpha$ and $\beta$ are set to be large in Fig.~\ref{fig:7}. 
The notable change in Fig.~\ref{fig:7} validates the importance of attention rectification and preservation components towards adversarial attacks. Fig.~\ref{fig:8} visualizes the activation map of two example images and their corresponding adversarial images w/ and w/o the proposed attention losses. The results justify our motivation to introduce the attention losses to rectify and preserve the attention area of both original and adversarial images. The best performance is generally obtained when setting $\alpha=\beta=0.01$. Around this ratio, the relatively stable accuracy curve shows that the proposed method is not very sensitive to the weight parameter configuration within a specific range.

\subsection{Attention Rectificaiton/Preservation in Existing Adversarial Training Solutions}

  \begin{figure}[t]
    \begin{minipage}[a]{1.0\linewidth}
        \centering
        \includegraphics[width=0.7\linewidth]{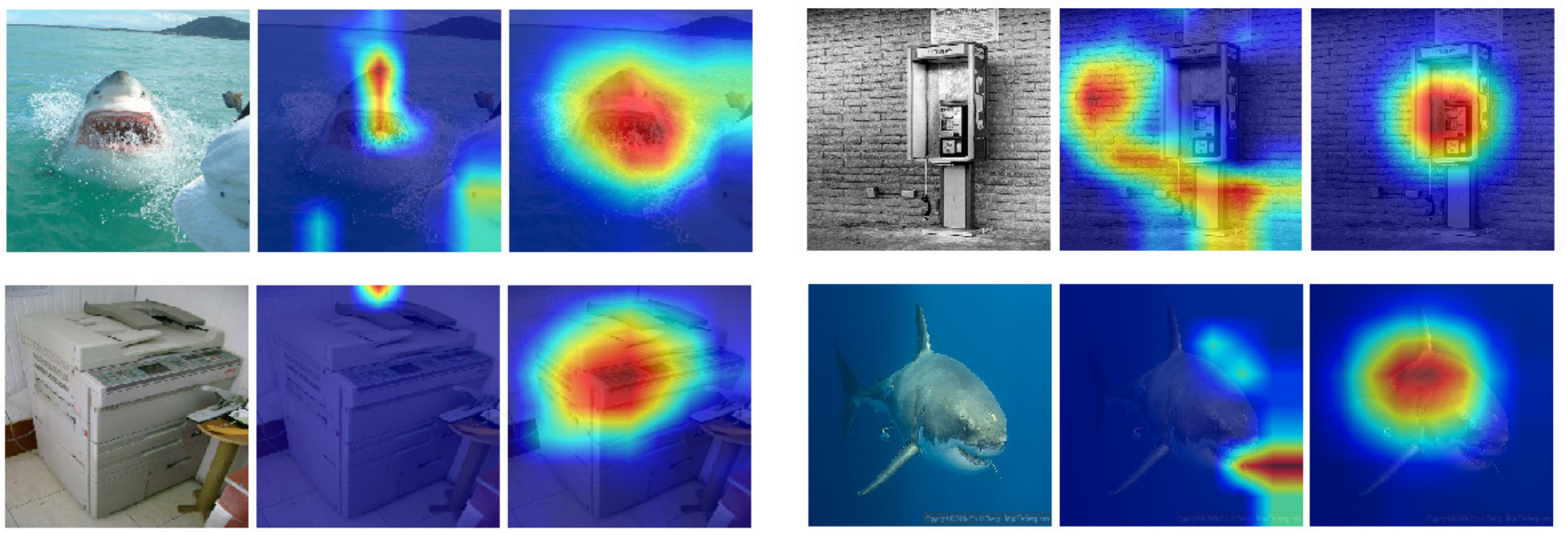}
        \centerline{(a)}
    \end{minipage}
    \begin{minipage}[b]{1.0\linewidth}
        \centering
        \includegraphics[width=0.7\linewidth]{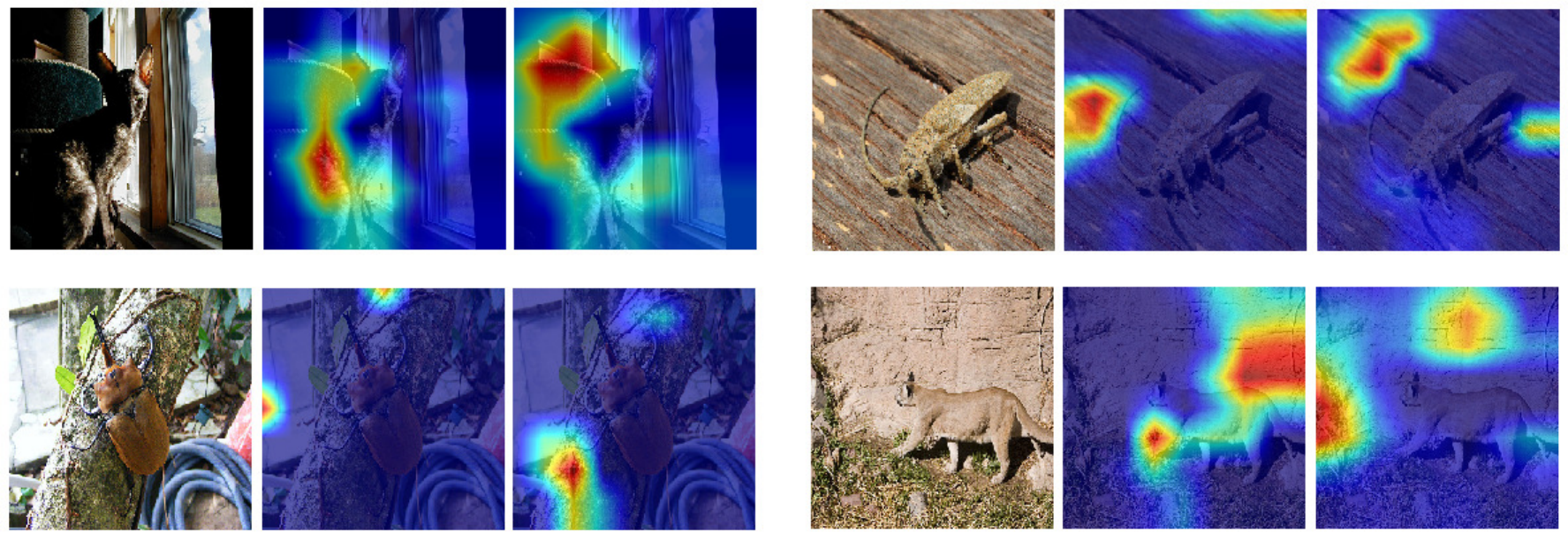}
        \centerline{(b)}
    \end{minipage}
    \caption{Example cases of (a) successful and (b) failed defense for adversarial training on ensemble of four attack methods~\cite{tramer2017ensemble}.  For each example, we show three images as the original image,  activation map of adversarial
    image for the clean model, and activation map of adversarial image for the adversarially trained model.}
    \label{fig:9}
\end{figure}

  \begin{figure}[tb]
    \begin{minipage}[a]{1.0\linewidth}
        \centering
        \includegraphics[width=0.7\linewidth]{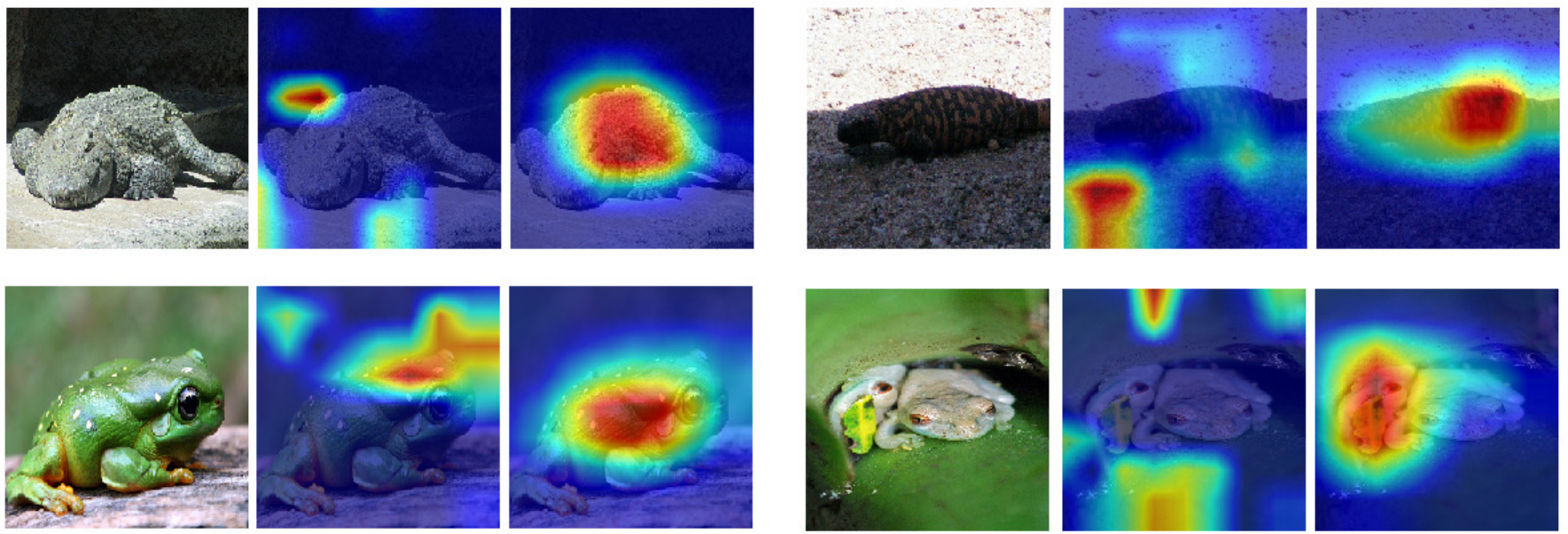}
        \centerline{(a)}
    \end{minipage}
    \begin{minipage}[b]{1.0\linewidth}
        \centering
        \includegraphics[width=0.7\linewidth]{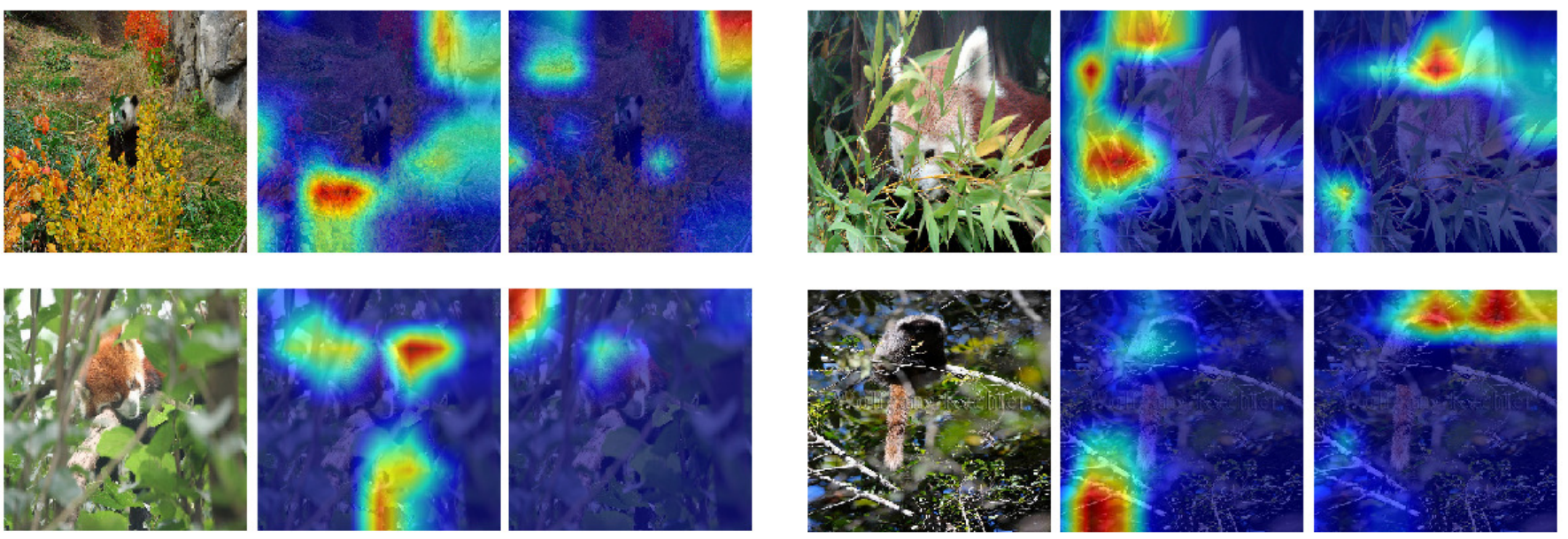}
        \centerline{(b)}
    \end{minipage}
    \caption{Example cases of (a) successful and (b) failed defense for adversarial training on 
    StepLL~\cite{kurakin2016adversarial}.  For each example, we show three images as the original image, activation map of adversarial
    image for the clean model, and activation map of adversarial image for the adversarially trained model.}
    \label{fig:a2}
\end{figure}

Inspired from the attention shrinkage and deviation observations, this study explicitly designs attention rectification and preservation solutions to improve adversarial robustness. By analyzing other existing adversarial defense solutions, we found that attention rectification/preservation is a common phenomenon to improve adversarial robustness.

Fig.~\ref{fig:9}(a) respectively illustrates the activation map change of four example images for adversarial training on an ensemble of four attack method~\cite{tramer2017ensemble}. It is shown that by adding adversarial images into the training set, the adversarial training method affects rectifying and preserving the activation maps of adversarial images. However, since standard adversarial training solutions have no explicit regularization concerning attention, attention rectification/preservation is not guaranteed in many cases.
Fig.~\ref{fig:9}(b) shows some examples where the examined adversarial training method fails to defend the adversarial attack. In these examples, no clear attention improvement is observed.

In order to prove the universality of the phenomenon, we also did the 
same analysis on the StepLL-based adversarial training method~\cite{tramer2017ensemble}. 
In the analysis of the StepLL-based adversarial training method, 
we can see that results are similar to the ensemble of four attack methods. Fig.~\ref{fig:a2}(a) shows that the adversarial training method affects rectifying and preserving the activation maps of adversarial images. Fig.~\ref{fig:a2}(b) shows some examples where the examined adversarial training method fails to defend the adversarial attack.

\subsection{Result of Attention-Augmented Adversarial Attack}
\begin{figure}[!t]
  \begin{center}
     \includegraphics[width=0.7\linewidth]{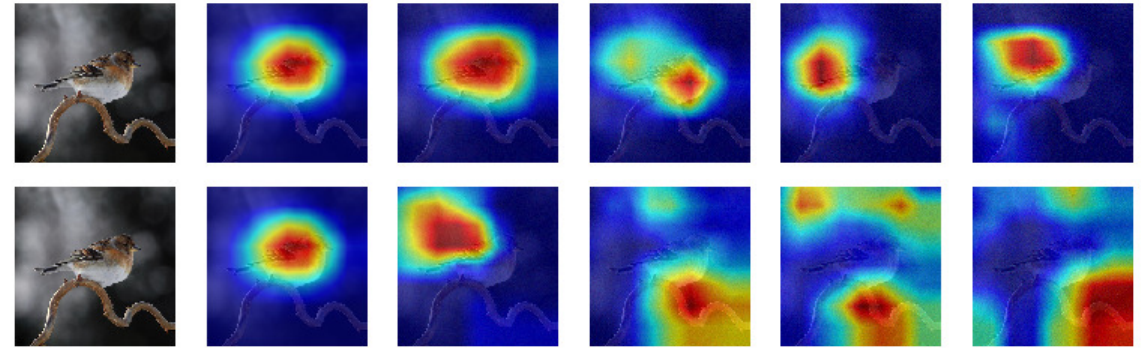}
  \end{center}
     \caption{Activation map change in different attack rounds under StepLL (top row) and StepLL+A$^3$Attack (bottom row).}
  \label{fig:12}
\end{figure}

\begin{table}[t]
    \begin{center}
    \begin{tabular}{|c|c|c|c|}
    \hline
    \multicolumn{2}{|c}{Iter-StepLL} & \multicolumn{2}{|c|}{PGD} \\ \hline
    Iter-StepLL & Iter-StepLL+A$^3$ & PDG & PGD+A$^3$ \\ \hline
    3.8\% & \textbf{3.3\%} & 13.6\% & \textbf{10.8\%} \\ \hline
    \end{tabular}
    \end{center}
    \caption{Adversarial attack classification accuracy on $\mathcal{I}_{att}$.}
    \label{tab:final}
\end{table}
  
In Sec.~\ref{AAA}, we use the constraint of distracting visual attention to propose an adversarial attack framework based on visual attention. Here we use preliminary experiments to prove that the adversarial attack using visual attention can play a role in the ordinary adversarial attack.

We conducted experiments on two widely used attack methods, Iter-StepLL and PGD, to validate the effectiveness of attention augmentation. Specifically, we use $\mathcal{I}_{att}$ as the dataset and select {$\sigma$}=10, $\epsilon$=4/255, iterations=3. Table~\ref{tab:final} shows the classification accuracy results. It is shown that considering additional attention to attack loss contributes to a stronger attack and more adversarial images are misclassified. 

In order to specifically observe how the Attention-Augmented Adversarial Attack framework improves the attack effect, we visualize the activation map of the entire attack process. Fig.~\ref{fig:12} illustrates the activation map change for one example image under Iter-StepLL and Iter-StepLL+A$^3$Attack. It is shown that the additional attention attack loss imposes an apparent effect to gradually distract attention from the original region, which increases the confidence to mislead the classifier. 

This preliminary result validates the effectiveness of considering the attention to adversarial image generation. The consideration of attention change provides a new way to investigate the problems of adversarial defense and attack. We hope that the results of this experiment can give some inspiration to those who study adversarial attacks.

\subsection{Analysis of Failure Cases}
\begin{figure}[t]
  \begin{minipage}{0.19\linewidth}
    \begin{center}
      \includegraphics[width=1.0\linewidth]{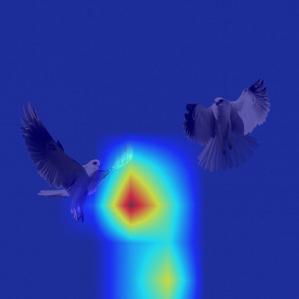}
   \end{center}
  \end{minipage}
  \begin{minipage}{0.19\linewidth}
    \begin{center}
      \includegraphics[width=1.0\linewidth]{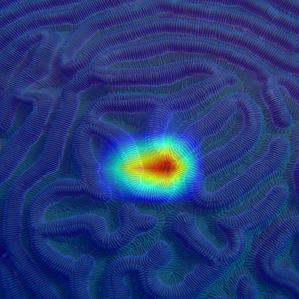}
   \end{center}
  \end{minipage}
  \begin{minipage}{0.19\linewidth}
    \begin{center}
      \includegraphics[width=1.0\linewidth]{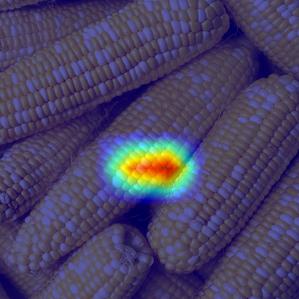}
   \end{center}
  \end{minipage}
  \begin{minipage}{0.19\linewidth}
    \begin{center}
      \includegraphics[width=1.0\linewidth]{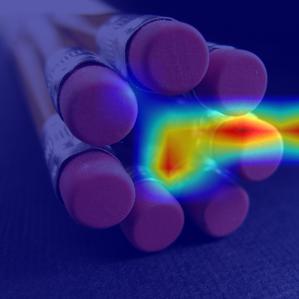}
   \end{center}
  \end{minipage}
  \begin{minipage}{0.19\linewidth}
    \begin{center}
      \includegraphics[width=1.0\linewidth]{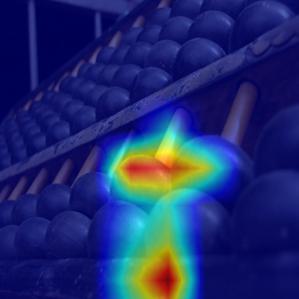}
   \end{center}
  \end{minipage}
  \caption{Some images that are difficult to form attention in the ImageNet2012 dataset.}
  \label{faild}
\end{figure}

During the experiments, we realized that the attention mechanism does not always work properly and fails on a small number of samples. Here we present this type of sample and analyze why it appears. The failed cases are shown in Fig.~\ref{faild}. We divide them into two categories:
\begin{itemize}
  \item The critical areas in the figure are too scattered, making it difficult to form a good attention map.
  \item Since irrelevant features in the dataset are too highly correlated with the target, the model relies too much on irrelevant features.
\end{itemize}

Since these two data types are relatively few in typical data sets, they will not significantly affect the algorithm's work. For this part of the data, although the attention module fails, the algorithm still performs adversarial training on it, so it can still defend against adversarial samples.
\section{Conclusions}

This study provides a new perspective to analyze the adversarial defense/attack problem by considering attention. Qualitative and quantitative experimental results demonstrate the effectiveness of attention-based adversarial defense/attack. In the future, we are working towards seeking insight into the mechanism behind the attention perturbations from adversarial attacks, as well as investigating other phenomena concerning attention observations (\emph{e.g.}, the scattered attention when adversarial attack proceeds) to inspire more comprehensive defense/attack solution.
Specifically, some recent studies~\cite{Zhang2019InterpretingAT} attribute the adversarial samples phenomenon to exploiting unsemantic features in derived models. We will explore why an accurate and stable activation map contributes to adversarial robustness from the perspective of semantic feature learning. We believe this will shed light on the correction between adversarial robustness and model interpretability, which coincides with the observation adversarial training helps improve model interpretability~\cite{Zhang2019InterpretingAT, Noack2019DoesIO}.

\section*{Acknowledgments}
This work is supported by the Fundamental Research Funds for the Central Universities (No. 2021JBM011).

\appendix

\bibliographystyle{ACM-Reference-Format}

\bibliography{acmart.bib}


\begin{thebibliography}{48}


\ifx \showCODEN    \undefined \def \showCODEN     #1{\unskip}     \fi
\ifx \showDOI      \undefined \def \showDOI       #1{#1}\fi
\ifx \showISBNx    \undefined \def \showISBNx     #1{\unskip}     \fi
\ifx \showISBNxiii \undefined \def \showISBNxiii  #1{\unskip}     \fi
\ifx \showISSN     \undefined \def \showISSN      #1{\unskip}     \fi
\ifx \showLCCN     \undefined \def \showLCCN      #1{\unskip}     \fi
\ifx \shownote     \undefined \def \shownote      #1{#1}          \fi
\ifx \showarticletitle \undefined \def \showarticletitle #1{#1}   \fi
\ifx \showURL      \undefined \def \showURL       {\relax}        \fi
\providecommand\bibfield[2]{#2}
\providecommand\bibinfo[2]{#2}
\providecommand\natexlab[1]{#1}
\providecommand\showeprint[2][]{arXiv:#2}

\bibitem[\protect\citeauthoryear{{Amini} and {Ghaemmaghami}}{{Amini} and
  {Ghaemmaghami}}{2020}]%
        {8970483}
\bibfield{author}{\bibinfo{person}{S. {Amini}} {and} \bibinfo{person}{S.
  {Ghaemmaghami}}.} \bibinfo{year}{2020}\natexlab{}.
\newblock \showarticletitle{Towards Improving Robustness of Deep Neural
  Networks to Adversarial Perturbations}.
\newblock \bibinfo{journal}{\emph{{IEEE} Transactions on Multimedia, {TMM}}}.
\newblock


\bibitem[\protect\citeauthoryear{Andriushchenko, Croce, Flammarion, and
  Hein}{Andriushchenko et~al\mbox{.}}{2020}]%
        {DBLP:conf/eccv/AndriushchenkoC20}
\bibfield{author}{\bibinfo{person}{Maksym Andriushchenko},
  \bibinfo{person}{Francesco Croce}, \bibinfo{person}{Nicolas Flammarion},
  {and} \bibinfo{person}{Matthias Hein}.} \bibinfo{year}{2020}\natexlab{}.
\newblock \showarticletitle{Square Attack: {A} Query-Efficient Black-Box
  Adversarial Attack via Random Search}. In \bibinfo{booktitle}{\emph{European
  Conference Computer Vision, {ECCV}}}.
\newblock


\bibitem[\protect\citeauthoryear{Athalye, Carlini, and Wagner}{Athalye
  et~al\mbox{.}}{2018}]%
        {athalye2018obfuscated}
\bibfield{author}{\bibinfo{person}{Anish Athalye}, \bibinfo{person}{Nicholas
  Carlini}, {and} \bibinfo{person}{David~A. Wagner}.}
  \bibinfo{year}{2018}\natexlab{}.
\newblock \showarticletitle{Obfuscated Gradients Give a False Sense of
  Security: Circumventing Defenses to Adversarial Examples}. In
  \bibinfo{booktitle}{\emph{International Conference on Machine Learning,
  {ICML}}}.
\newblock


\bibitem[\protect\citeauthoryear{Carlini and Wagner}{Carlini and
  Wagner}{2017}]%
        {nicholas2017towards}
\bibfield{author}{\bibinfo{person}{Nicholas Carlini} {and}
  \bibinfo{person}{David~A. Wagner}.} \bibinfo{year}{2017}\natexlab{}.
\newblock \showarticletitle{Towards Evaluating the Robustness of Neural
  Networks}. In \bibinfo{booktitle}{\emph{{IEEE} Symposium on Security and
  Privacy, {SP}}}.
\newblock


\bibitem[\protect\citeauthoryear{Croce and Hein}{Croce and Hein}{2020}]%
        {DBLP:conf/icml/Croce020a}
\bibfield{author}{\bibinfo{person}{Francesco Croce} {and}
  \bibinfo{person}{Matthias Hein}.} \bibinfo{year}{2020}\natexlab{}.
\newblock \showarticletitle{Reliable evaluation of adversarial robustness with
  an ensemble of diverse parameter-free attacks}. In
  \bibinfo{booktitle}{\emph{International Conference on Machine Learning,
  {ICML}}}.
\newblock


\bibitem[\protect\citeauthoryear{Deng, Dong, Socher, Li, Li, and Li}{Deng
  et~al\mbox{.}}{2009}]%
        {deng2009imagenet}
\bibfield{author}{\bibinfo{person}{Jia Deng}, \bibinfo{person}{Wei Dong},
  \bibinfo{person}{Richard Socher}, \bibinfo{person}{Li{-}Jia Li},
  \bibinfo{person}{Kai Li}, {and} \bibinfo{person}{Fei{-}Fei Li}.}
  \bibinfo{year}{2009}\natexlab{}.
\newblock \showarticletitle{ImageNet: {A} large-scale hierarchical image
  database}. In \bibinfo{booktitle}{\emph{{IEEE} Computer Society Conference on
  Computer Vision and Pattern Recognition, {CVPR}}}.
\newblock


\bibitem[\protect\citeauthoryear{Dhillon, Azizzadenesheli, Lipton, Bernstein,
  Kossaifi, Khanna, and Anandkumar}{Dhillon et~al\mbox{.}}{2018}]%
        {dhillon2018stochastic}
\bibfield{author}{\bibinfo{person}{Guneet~S. Dhillon}, \bibinfo{person}{Kamyar
  Azizzadenesheli}, \bibinfo{person}{Zachary~C. Lipton},
  \bibinfo{person}{Jeremy Bernstein}, \bibinfo{person}{Jean Kossaifi},
  \bibinfo{person}{Aran Khanna}, {and} \bibinfo{person}{Anima Anandkumar}.}
  \bibinfo{year}{2018}\natexlab{}.
\newblock \showarticletitle{Stochastic Activation Pruning for Robust
  Adversarial Defense}.
\newblock \bibinfo{journal}{\emph{CoRR}}.
\newblock


\bibitem[\protect\citeauthoryear{Dong, Fu, Yang, Pang, Su, Xiao, and Zhu}{Dong
  et~al\mbox{.}}{2020}]%
        {DBLP:conf/cvpr/DongFYPSXZ20}
\bibfield{author}{\bibinfo{person}{Yinpeng Dong}, \bibinfo{person}{Qi{-}An Fu},
  \bibinfo{person}{Xiao Yang}, \bibinfo{person}{Tianyu Pang},
  \bibinfo{person}{Hang Su}, \bibinfo{person}{Zihao Xiao}, {and}
  \bibinfo{person}{Jun Zhu}.} \bibinfo{year}{2020}\natexlab{}.
\newblock \showarticletitle{Benchmarking Adversarial Robustness on Image
  Classification}. In \bibinfo{booktitle}{\emph{{IEEE} Conference on Computer
  Vision and Pattern Recognition, {CVPR}}}.
\newblock


\bibitem[\protect\citeauthoryear{{Du}, {Fang}, {Yi}, {Xu}, {Cheng}, and
  {Tao}}{{Du} et~al\mbox{.}}{2019}]%
        {8576563}
\bibfield{author}{\bibinfo{person}{Y. {Du}}, \bibinfo{person}{M. {Fang}},
  \bibinfo{person}{J. {Yi}}, \bibinfo{person}{C. {Xu}}, \bibinfo{person}{J.
  {Cheng}}, {and} \bibinfo{person}{D. {Tao}}.} \bibinfo{year}{2019}\natexlab{}.
\newblock \showarticletitle{Enhancing the Robustness of Neural Collaborative
  Filtering Systems Under Malicious Attacks}.
\newblock \bibinfo{journal}{\emph{{IEEE} Transactions on Multimedia, {TMM}}}.
\newblock


\bibitem[\protect\citeauthoryear{Duan, Li, Deng, Xiao, and Tian}{Duan
  et~al\mbox{.}}{2022}]%
        {DBLP:journals/tomccap/DuanLDXT22}
\bibfield{author}{\bibinfo{person}{Mingxing Duan}, \bibinfo{person}{Kenli Li},
  \bibinfo{person}{Jiayan Deng}, \bibinfo{person}{Bin Xiao}, {and}
  \bibinfo{person}{Qi Tian}.} \bibinfo{year}{2022}\natexlab{}.
\newblock \showarticletitle{A Novel Multi-Sample Generation Method for
  Adversarial Attacks}.
\newblock \bibinfo{journal}{\emph{{ACM} Trans. Multim. Comput. Commun. Appl.,
  {ToMM}}}.
\newblock


\bibitem[\protect\citeauthoryear{Duan, Li, Ouyang, Win, Li, and Tian}{Duan
  et~al\mbox{.}}{2020}]%
        {DBLP:journals/tomccap/DuanLOWLT20}
\bibfield{author}{\bibinfo{person}{Mingxing Duan}, \bibinfo{person}{Kenli Li},
  \bibinfo{person}{Aijia Ouyang}, \bibinfo{person}{Khin~Nandar Win},
  \bibinfo{person}{Keqin Li}, {and} \bibinfo{person}{Qi Tian}.}
  \bibinfo{year}{2020}\natexlab{}.
\newblock \showarticletitle{EGroupNet: {A} Feature-enhanced Network for Age
  Estimation with Novel Age Group Schemes}.
\newblock \bibinfo{journal}{\emph{{ACM} Trans. Multim. Comput. Commun. Appl.,
  {ToMM}}}.
\newblock


\bibitem[\protect\citeauthoryear{Duan, Li, Xie, Tian, and Xiao}{Duan
  et~al\mbox{.}}{2021}]%
        {DBLP:conf/mm/DuanLX0X21}
\bibfield{author}{\bibinfo{person}{Mingxing Duan}, \bibinfo{person}{Kenli Li},
  \bibinfo{person}{Lingxi Xie}, \bibinfo{person}{Qi Tian}, {and}
  \bibinfo{person}{Bin Xiao}.} \bibinfo{year}{2021}\natexlab{}.
\newblock \showarticletitle{Towards Multiple Black-boxes Attack via Adversarial
  Example Generation Network}. In \bibinfo{booktitle}{\emph{{ACM} Multimedia
  Conference, {MM}}}.
\newblock


\bibitem[\protect\citeauthoryear{Ferrari, Becattini, Galteri, and
  Bimbo}{Ferrari et~al\mbox{.}}{2022}]%
        {10.1145/3524619}
\bibfield{author}{\bibinfo{person}{Claudio Ferrari}, \bibinfo{person}{Federico
  Becattini}, \bibinfo{person}{Leonardo Galteri}, {and}
  \bibinfo{person}{Alberto~Del Bimbo}.} \bibinfo{year}{2022}\natexlab{}.
\newblock \showarticletitle{(Compress and Restore)N: A Robust Defense Against
  Adversarial Attacks on Image Classification}.
\newblock \bibinfo{journal}{\emph{{ACM} Trans. Multimedia Comput. Commun.
  Appl., {ToMM}}}.
\newblock


\bibitem[\protect\citeauthoryear{Goodfellow, Shlens, and Szegedy}{Goodfellow
  et~al\mbox{.}}{2015}]%
        {Ian2017Explaining}
\bibfield{author}{\bibinfo{person}{Ian Goodfellow}, \bibinfo{person}{Jonathon
  Shlens}, {and} \bibinfo{person}{Christian Szegedy}.}
  \bibinfo{year}{2015}\natexlab{}.
\newblock \showarticletitle{Explaining and Harnessing Adversarial Examples}. In
  \bibinfo{booktitle}{\emph{International Conference on Learning
  Representations, {ICLR}}}.
\newblock


\bibitem[\protect\citeauthoryear{Guo, Rana, Ciss{\'{e}}, and van~der
  Maaten}{Guo et~al\mbox{.}}{2017}]%
        {guo2017countering}
\bibfield{author}{\bibinfo{person}{Chuan Guo}, \bibinfo{person}{Mayank Rana},
  \bibinfo{person}{Moustapha Ciss{\'{e}}}, {and} \bibinfo{person}{Laurens
  van~der Maaten}.} \bibinfo{year}{2017}\natexlab{}.
\newblock \showarticletitle{Countering Adversarial Images using Input
  Transformations}.
\newblock \bibinfo{journal}{\emph{CoRR}}.
\newblock


\bibitem[\protect\citeauthoryear{He, Zhang, Ren, and Sun}{He
  et~al\mbox{.}}{2016}]%
        {resnet}
\bibfield{author}{\bibinfo{person}{Kaiming He}, \bibinfo{person}{Xiangyu
  Zhang}, \bibinfo{person}{Shaoqing Ren}, {and} \bibinfo{person}{Jian Sun}.}
  \bibinfo{year}{2016}\natexlab{}.
\newblock \showarticletitle{Deep Residual Learning for Image Recognition}. In
  \bibinfo{booktitle}{\emph{{IEEE} Conference on Computer Vision and Pattern
  Recognition, {CVPR}}}.
\newblock


\bibitem[\protect\citeauthoryear{Hinton, Vinyals, and Dean}{Hinton
  et~al\mbox{.}}{2015}]%
        {hinton2015distilling}
\bibfield{author}{\bibinfo{person}{Geoffrey~E. Hinton}, \bibinfo{person}{Oriol
  Vinyals}, {and} \bibinfo{person}{Jeffrey Dean}.}
  \bibinfo{year}{2015}\natexlab{}.
\newblock \showarticletitle{Distilling the Knowledge in a Neural Network}.
\newblock \bibinfo{journal}{\emph{CoRR}}.
\newblock


\bibitem[\protect\citeauthoryear{Kannan, Kurakin, and Goodfellow}{Kannan
  et~al\mbox{.}}{2018}]%
        {alp}
\bibfield{author}{\bibinfo{person}{Harini Kannan}, \bibinfo{person}{Alexey
  Kurakin}, {and} \bibinfo{person}{Ian~J. Goodfellow}.}
  \bibinfo{year}{2018}\natexlab{}.
\newblock \showarticletitle{Adversarial Logit Pairing}.
\newblock \bibinfo{journal}{\emph{CoRR}}.
\newblock


\bibitem[\protect\citeauthoryear{Krizhevsky and Hinton}{Krizhevsky and
  Hinton}{2009}]%
        {krizhevsky2009learning}
\bibfield{author}{\bibinfo{person}{Alex Krizhevsky} {and}
  \bibinfo{person}{Geoffrey Hinton}.} \bibinfo{year}{2009}\natexlab{}.
\newblock \showarticletitle{Learning multiple layers of features from tiny
  images}. In \bibinfo{booktitle}{\emph{Citeseer}}.
\newblock


\bibitem[\protect\citeauthoryear{Kurakin, Goodfellow, and Bengio}{Kurakin
  et~al\mbox{.}}{2016}]%
        {kurakin2016adversarial}
\bibfield{author}{\bibinfo{person}{Alexey Kurakin}, \bibinfo{person}{Ian~J.
  Goodfellow}, {and} \bibinfo{person}{Samy Bengio}.}
  \bibinfo{year}{2016}\natexlab{}.
\newblock \showarticletitle{Adversarial examples in the physical world}. In
  \bibinfo{booktitle}{\emph{International Conference on Learning
  Representations, {ICLR}}}.
\newblock


\bibitem[\protect\citeauthoryear{Kurakin, Goodfellow, and Bengio}{Kurakin
  et~al\mbox{.}}{2017}]%
        {KurakinGB17}
\bibfield{author}{\bibinfo{person}{Alexey Kurakin}, \bibinfo{person}{Ian~J.
  Goodfellow}, {and} \bibinfo{person}{Samy Bengio}.}
  \bibinfo{year}{2017}\natexlab{}.
\newblock \showarticletitle{Adversarial Machine Learning at Scale}. In
  \bibinfo{booktitle}{\emph{International Conference on Learning
  Representations, {ICLR}}}.
\newblock


\bibitem[\protect\citeauthoryear{{Li}, {Zeng}, {Li}, {Lin}, and {Yu}}{{Li}
  et~al\mbox{.}}{2020}]%
        {9206141}
\bibfield{author}{\bibinfo{person}{H. {Li}}, \bibinfo{person}{Y. {Zeng}},
  \bibinfo{person}{G. {Li}}, \bibinfo{person}{L. {Lin}}, {and}
  \bibinfo{person}{Y. {Yu}}.} \bibinfo{year}{2020}\natexlab{}.
\newblock \showarticletitle{Online Alternate Generator Against Adversarial
  Attacks}.
\newblock \bibinfo{journal}{\emph{{IEEE} Transactions on Image Processing,
  {TIP}}}.
\newblock


\bibitem[\protect\citeauthoryear{Liao, Liang, Dong, Pang, Hu, and Zhu}{Liao
  et~al\mbox{.}}{2018a}]%
        {Fangzhouy2017Defense}
\bibfield{author}{\bibinfo{person}{Fangzhou Liao}, \bibinfo{person}{Ming
  Liang}, \bibinfo{person}{Yinpeng Dong}, \bibinfo{person}{Tianyu Pang},
  \bibinfo{person}{Xiaolin Hu}, {and} \bibinfo{person}{Jun Zhu}.}
  \bibinfo{year}{2018}\natexlab{a}.
\newblock \showarticletitle{Defense Against Adversarial Attacks Using
  High-Level Representation Guided Denoiser}. In
  \bibinfo{booktitle}{\emph{{IEEE} Conference on Computer Vision and Pattern
  Recognition, {CVPR}}}.
\newblock


\bibitem[\protect\citeauthoryear{Liao, Liang, Dong, Pang, Hu, and Zhu}{Liao
  et~al\mbox{.}}{2018b}]%
        {liao2017defense}
\bibfield{author}{\bibinfo{person}{Fangzhou Liao}, \bibinfo{person}{Ming
  Liang}, \bibinfo{person}{Yinpeng Dong}, \bibinfo{person}{Tianyu Pang},
  \bibinfo{person}{Xiaolin Hu}, {and} \bibinfo{person}{Jun Zhu}.}
  \bibinfo{year}{2018}\natexlab{b}.
\newblock \showarticletitle{Defense Against Adversarial Attacks Using
  High-Level Representation Guided Denoiser}. In
  \bibinfo{booktitle}{\emph{{IEEE} Conference on Computer Vision and Pattern
  Recognition, {CVPR}}}.
\newblock


\bibitem[\protect\citeauthoryear{Liu and J{\'{a}}J{\'{a}}}{Liu and
  J{\'{a}}J{\'{a}}}{2018}]%
        {liu2018feature}
\bibfield{author}{\bibinfo{person}{Chihuang Liu} {and} \bibinfo{person}{Joseph
  J{\'{a}}J{\'{a}}}.} \bibinfo{year}{2018}\natexlab{}.
\newblock \showarticletitle{Feature prioritization and regularization improve
  standard accuracy and adversarial robustness}.
\newblock \bibinfo{journal}{\emph{CoRR}}.
\newblock


\bibitem[\protect\citeauthoryear{{Liu}, {Liu}, {Zhang}, {Liu}, and
  {Shen}}{{Liu} et~al\mbox{.}}{2021}]%
        {9335499}
\bibfield{author}{\bibinfo{person}{F. {Liu}}, \bibinfo{person}{H. {Liu}},
  \bibinfo{person}{W. {Zhang}}, \bibinfo{person}{G. {Liu}}, {and}
  \bibinfo{person}{L. {Shen}}.} \bibinfo{year}{2021}\natexlab{}.
\newblock \showarticletitle{One-Class Fingerprint Presentation Attack Detection
  Using Auto-Encoder Network}.
\newblock \bibinfo{journal}{\emph{{IEEE} Transactions on Image Processing,
  {TIP}}}.
\newblock


\bibitem[\protect\citeauthoryear{Madry, Makelov, Schmidt, Tsipras, and
  Vladu}{Madry et~al\mbox{.}}{2017}]%
        {madry2017towards}
\bibfield{author}{\bibinfo{person}{Aleksander Madry},
  \bibinfo{person}{Aleksandar Makelov}, \bibinfo{person}{Ludwig Schmidt},
  \bibinfo{person}{Dimitris Tsipras}, {and} \bibinfo{person}{Adrian Vladu}.}
  \bibinfo{year}{2017}\natexlab{}.
\newblock \showarticletitle{Towards Deep Learning Models Resistant to
  Adversarial Attacks}. In \bibinfo{booktitle}{\emph{International Conference
  on Learning Representations, {ICLR}}}.
\newblock


\bibitem[\protect\citeauthoryear{{Mohajerani} and {Saeedi}}{{Mohajerani} and
  {Saeedi}}{2019}]%
        {8664462}
\bibfield{author}{\bibinfo{person}{S. {Mohajerani}} {and} \bibinfo{person}{P.
  {Saeedi}}.} \bibinfo{year}{2019}\natexlab{}.
\newblock \showarticletitle{Shadow Detection in Single RGB Images Using a
  Context Preserver Convolutional Neural Network Trained by Multiple
  Adversarial Examples}.
\newblock \bibinfo{journal}{\emph{{IEEE} Transactions on Image Processing,
  {TIP}}}.
\newblock


\bibitem[\protect\citeauthoryear{Moosavi{-}Dezfooli, Fawzi, and
  Frossard}{Moosavi{-}Dezfooli et~al\mbox{.}}{2016}]%
        {Seyed-Mohsen2016Deepfool}
\bibfield{author}{\bibinfo{person}{Seyed{-}Mohsen Moosavi{-}Dezfooli},
  \bibinfo{person}{Alhussein Fawzi}, {and} \bibinfo{person}{Pascal Frossard}.}
  \bibinfo{year}{2016}\natexlab{}.
\newblock \showarticletitle{DeepFool: {A} Simple and Accurate Method to Fool
  Deep Neural Networks}. In \bibinfo{booktitle}{\emph{{IEEE} Conference on
  Computer Vision and Pattern Recognition, {CVPR}}}.
\newblock


\bibitem[\protect\citeauthoryear{{Mustafa}, {Khan}, {Hayat}, {Shen}, and
  {Shao}}{{Mustafa} et~al\mbox{.}}{2020}]%
        {8844865}
\bibfield{author}{\bibinfo{person}{A. {Mustafa}}, \bibinfo{person}{S.~H.
  {Khan}}, \bibinfo{person}{M. {Hayat}}, \bibinfo{person}{J. {Shen}}, {and}
  \bibinfo{person}{L. {Shao}}.} \bibinfo{year}{2020}\natexlab{}.
\newblock \showarticletitle{Image Super-Resolution as a Defense Against
  Adversarial Attacks}.
\newblock \bibinfo{journal}{\emph{{IEEE} Transactions on Image Processing,
  {TIP}}}.
\newblock


\bibitem[\protect\citeauthoryear{Noack, Ahern, Dou, and Li}{Noack
  et~al\mbox{.}}{2019}]%
        {Noack2019DoesIO}
\bibfield{author}{\bibinfo{person}{A. Noack}, \bibinfo{person}{Isaac Ahern},
  \bibinfo{person}{D. Dou}, {and} \bibinfo{person}{Boyang Li}.}
  \bibinfo{year}{2019}\natexlab{}.
\newblock \showarticletitle{Does Interpretability of Neural Networks Imply
  Adversarial Robustness?}
\newblock \bibinfo{journal}{\emph{CoRR}}.
\newblock


\bibitem[\protect\citeauthoryear{Papernot, McDaniel, Wu, Jha, and
  Swami}{Papernot et~al\mbox{.}}{2016}]%
        {papernot2016distillation}
\bibfield{author}{\bibinfo{person}{Nicolas Papernot},
  \bibinfo{person}{Patrick~D. McDaniel}, \bibinfo{person}{Xi Wu},
  \bibinfo{person}{Somesh Jha}, {and} \bibinfo{person}{Ananthram Swami}.}
  \bibinfo{year}{2016}\natexlab{}.
\newblock \showarticletitle{Distillation as a Defense to Adversarial
  Perturbations Against Deep Neural Networks}. In
  \bibinfo{booktitle}{\emph{{IEEE} Symposium on Security and Privacy, {SP}}}.
\newblock


\bibitem[\protect\citeauthoryear{{Raghavendra}, {Raja}, and
  {Busch}}{{Raghavendra} et~al\mbox{.}}{2015}]%
        {7018027}
\bibfield{author}{\bibinfo{person}{R. {Raghavendra}}, \bibinfo{person}{K.~B.
  {Raja}}, {and} \bibinfo{person}{C. {Busch}}.}
  \bibinfo{year}{2015}\natexlab{}.
\newblock \showarticletitle{Presentation Attack Detection for Face Recognition
  Using Light Field Camera}.
\newblock \bibinfo{journal}{\emph{{IEEE} Transactions on Image Processing,
  {TIP}}}.
\newblock


\bibitem[\protect\citeauthoryear{Ribeiro, Singh, and Guestrin}{Ribeiro
  et~al\mbox{.}}{2016}]%
        {ribeiro2016should}
\bibfield{author}{\bibinfo{person}{Marco~T{\'{u}}lio Ribeiro},
  \bibinfo{person}{Sameer Singh}, {and} \bibinfo{person}{Carlos Guestrin}.}
  \bibinfo{year}{2016}\natexlab{}.
\newblock \showarticletitle{"Why Should {I} Trust You?": Explaining the
  Predictions of Any Classifier}. In \bibinfo{booktitle}{\emph{{ACM}
  International Conference on Knowledge Discovery and Data Mining, {SIGKDD}}}.
\newblock


\bibitem[\protect\citeauthoryear{Selvaraju, Cogswell, Das, Vedantam, Parikh,
  and Batra}{Selvaraju et~al\mbox{.}}{2017}]%
        {selvaraju2017grad}
\bibfield{author}{\bibinfo{person}{Ramprasaath~R. Selvaraju},
  \bibinfo{person}{Michael Cogswell}, \bibinfo{person}{Abhishek Das},
  \bibinfo{person}{Ramakrishna Vedantam}, \bibinfo{person}{Devi Parikh}, {and}
  \bibinfo{person}{Dhruv Batra}.} \bibinfo{year}{2017}\natexlab{}.
\newblock \showarticletitle{Grad-CAM: Visual Explanations from Deep Networks
  via Gradient-Based Localization}. In \bibinfo{booktitle}{\emph{{IEEE}
  International Conference on Computer Vision, {ICCV}}}.
\newblock


\bibitem[\protect\citeauthoryear{Shen, Jin, Gao, and Zhang}{Shen
  et~al\mbox{.}}{2017}]%
        {shen2017ape}
\bibfield{author}{\bibinfo{person}{Shiwei Shen}, \bibinfo{person}{Guoqing Jin},
  \bibinfo{person}{Ke Gao}, {and} \bibinfo{person}{Yongdong Zhang}.}
  \bibinfo{year}{2017}\natexlab{}.
\newblock \showarticletitle{Ape-gan: Adversarial perturbation elimination with
  gan}.
\newblock \bibinfo{journal}{\emph{CoRR}}.
\newblock


\bibitem[\protect\citeauthoryear{Song, Kim, Nowozin, Ermon, and Kushman}{Song
  et~al\mbox{.}}{2017}]%
        {song2017pixeldefend}
\bibfield{author}{\bibinfo{person}{Yang Song}, \bibinfo{person}{Taesup Kim},
  \bibinfo{person}{Sebastian Nowozin}, \bibinfo{person}{Stefano Ermon}, {and}
  \bibinfo{person}{Nate Kushman}.} \bibinfo{year}{2017}\natexlab{}.
\newblock \showarticletitle{PixelDefend: Leveraging Generative Models to
  Understand and Defend against Adversarial Examples}.
\newblock \bibinfo{journal}{\emph{CoRR}}.
\newblock


\bibitem[\protect\citeauthoryear{{Su}, {Fang}, {Wang}, {Mehrotra}, {Begen},
  {Ye}, and {Cavallaro}}{{Su} et~al\mbox{.}}{2019}]%
        {8649865}
\bibfield{author}{\bibinfo{person}{Z. {Su}}, \bibinfo{person}{Q. {Fang}},
  \bibinfo{person}{H. {Wang}}, \bibinfo{person}{S. {Mehrotra}},
  \bibinfo{person}{A.~C. {Begen}}, \bibinfo{person}{Q. {Ye}}, {and}
  \bibinfo{person}{A. {Cavallaro}}.} \bibinfo{year}{2019}\natexlab{}.
\newblock \showarticletitle{Guest Editorial Trustworthiness in Social
  Multimedia Analytics and Delivery}.
\newblock \bibinfo{journal}{\emph{{IEEE} Transactions on Multimedia, {TMM}}}.
\newblock


\bibitem[\protect\citeauthoryear{Szegedy, Vanhoucke, Ioffe, Shlens, and
  Wojna}{Szegedy et~al\mbox{.}}{2016}]%
        {szegedy2016rethinking}
\bibfield{author}{\bibinfo{person}{Christian Szegedy}, \bibinfo{person}{Vincent
  Vanhoucke}, \bibinfo{person}{Sergey Ioffe}, \bibinfo{person}{Jonathon
  Shlens}, {and} \bibinfo{person}{Zbigniew Wojna}.}
  \bibinfo{year}{2016}\natexlab{}.
\newblock \showarticletitle{Rethinking the Inception Architecture for Computer
  Vision}. In \bibinfo{booktitle}{\emph{{IEEE} Conference on Computer Vision
  and Pattern Recognition, {CVPR}}}.
\newblock


\bibitem[\protect\citeauthoryear{Szegedy, Zaremba, Sutskever, Bruna, Erhan,
  Goodfellow, and Fergus}{Szegedy et~al\mbox{.}}{2014}]%
        {szegedy2013intriguing}
\bibfield{author}{\bibinfo{person}{Christian Szegedy},
  \bibinfo{person}{Wojciech Zaremba}, \bibinfo{person}{Ilya Sutskever},
  \bibinfo{person}{Joan Bruna}, \bibinfo{person}{Dumitru Erhan},
  \bibinfo{person}{Ian Goodfellow}, {and} \bibinfo{person}{Rob Fergus}.}
  \bibinfo{year}{2014}\natexlab{}.
\newblock \showarticletitle{Intriguing properties of neural networks}. In
  \bibinfo{booktitle}{\emph{International Conference on Learning
  Representations, {ICLR}}}.
\newblock


\bibitem[\protect\citeauthoryear{Tram{\`{e}}r, Kurakin, Papernot, Boneh, and
  McDaniel}{Tram{\`{e}}r et~al\mbox{.}}{2017}]%
        {tramer2017ensemble}
\bibfield{author}{\bibinfo{person}{Florian Tram{\`{e}}r},
  \bibinfo{person}{Alexey Kurakin}, \bibinfo{person}{Nicolas Papernot},
  \bibinfo{person}{Dan Boneh}, {and} \bibinfo{person}{Patrick~D. McDaniel}.}
  \bibinfo{year}{2017}\natexlab{}.
\newblock \showarticletitle{Ensemble Adversarial Training: Attacks and
  Defenses}. In \bibinfo{booktitle}{\emph{International Conference on Learning
  Representations, {ICLR}}}.
\newblock


\bibitem[\protect\citeauthoryear{Tsipras, Santurkar, Engstrom, Turner, and
  Madry}{Tsipras et~al\mbox{.}}{2018}]%
        {tsipras2018there}
\bibfield{author}{\bibinfo{person}{Dimitris Tsipras}, \bibinfo{person}{Shibani
  Santurkar}, \bibinfo{person}{Logan Engstrom}, \bibinfo{person}{Alexander
  Turner}, {and} \bibinfo{person}{Aleksander Madry}.}
  \bibinfo{year}{2018}\natexlab{}.
\newblock \showarticletitle{There Is No Free Lunch In Adversarial Robustness
  (But There Are Unexpected Benefits)}.
\newblock \bibinfo{journal}{\emph{CoRR}}.
\newblock


\bibitem[\protect\citeauthoryear{{Wang}, {Su}, {Zhang}, and {Hu}}{{Wang}
  et~al\mbox{.}}{2019}]%
        {8884184}
\bibfield{author}{\bibinfo{person}{Y. {Wang}}, \bibinfo{person}{H. {Su}},
  \bibinfo{person}{B. {Zhang}}, {and} \bibinfo{person}{X. {Hu}}.}
  \bibinfo{year}{2019}\natexlab{}.
\newblock \showarticletitle{Learning Reliable Visual Saliency for Model
  Explanations}.
\newblock \bibinfo{journal}{\emph{{IEEE} Transactions on Multimedia, {TMM}}}.
\newblock


\bibitem[\protect\citeauthoryear{Xiao and Zheng}{Xiao and Zheng}{2020}]%
        {DBLP:conf/cvpr/XiaoZ20}
\bibfield{author}{\bibinfo{person}{Chang Xiao} {and} \bibinfo{person}{Changxi
  Zheng}.} \bibinfo{year}{2020}\natexlab{}.
\newblock \showarticletitle{One Man's Trash Is Another Man's Treasure:
  Resisting Adversarial Examples by Adversarial Examples}. In
  \bibinfo{booktitle}{\emph{{IEEE} Conference on Computer Vision and Pattern
  Recognition, {CVPR}}}.
\newblock


\bibitem[\protect\citeauthoryear{Zeiler and Fergus}{Zeiler and Fergus}{2014}]%
        {zeiler2014visualizing}
\bibfield{author}{\bibinfo{person}{Matthew~D. Zeiler} {and}
  \bibinfo{person}{Rob Fergus}.} \bibinfo{year}{2014}\natexlab{}.
\newblock \showarticletitle{Visualizing and Understanding Convolutional
  Networks}. In \bibinfo{booktitle}{\emph{European Conference Computer Vision,
  {ECCV}}}.
\newblock


\bibitem[\protect\citeauthoryear{{Zhang}, {Liu}, {Liu}, {Xu}, {Yu}, {Ma}, and
  {Li}}{{Zhang} et~al\mbox{.}}{2021}]%
        {9286885}
\bibfield{author}{\bibinfo{person}{C. {Zhang}}, \bibinfo{person}{A. {Liu}},
  \bibinfo{person}{X. {Liu}}, \bibinfo{person}{Y. {Xu}}, \bibinfo{person}{H.
  {Yu}}, \bibinfo{person}{Y. {Ma}}, {and} \bibinfo{person}{T. {Li}}.}
  \bibinfo{year}{2021}\natexlab{}.
\newblock \showarticletitle{Interpreting and Improving Adversarial Robustness
  of Deep Neural Networks With Neuron Sensitivity}.
\newblock \bibinfo{journal}{\emph{{IEEE} Transactions on Image Processing,
  {TIP}}}.
\newblock


\bibitem[\protect\citeauthoryear{Zhang, Yu, Jiao, Xing, Ghaoui, and
  Jordan}{Zhang et~al\mbox{.}}{2019}]%
        {pmlr-v97-zhang19p}
\bibfield{author}{\bibinfo{person}{Hongyang Zhang}, \bibinfo{person}{Yaodong
  Yu}, \bibinfo{person}{Jiantao Jiao}, \bibinfo{person}{Eric Xing},
  \bibinfo{person}{Laurent~El Ghaoui}, {and} \bibinfo{person}{Michael Jordan}.}
  \bibinfo{year}{2019}\natexlab{}.
\newblock \showarticletitle{Theoretically Principled Trade-off between
  Robustness and Accuracy}. In \bibinfo{booktitle}{\emph{International
  Conference on Machine Learning, {ICML}}}.
\newblock


\bibitem[\protect\citeauthoryear{Zhang and Zhu}{Zhang and Zhu}{2019}]%
        {Zhang2019InterpretingAT}
\bibfield{author}{\bibinfo{person}{Tianyuan Zhang} {and}
  \bibinfo{person}{Zhanxing Zhu}.} \bibinfo{year}{2019}\natexlab{}.
\newblock \showarticletitle{Interpreting Adversarially Trained Convolutional
  Neural Networks}.
\newblock \bibinfo{journal}{\emph{CoRR}}.
\newblock


\end{thebibliography}

\end{document}